\documentclass[10pt]{article} 

\usepackage[preprint]{tmlr}

\usepackage{amsmath,amsfonts,bm}

\def\eqref#1{equation~\ref{#1}}

\def\1{\bm{1}}

\def\vp{{\bm{p}}}
\def\vq{{\bm{q}}}

\def\vx{{\bm{x}}}

\def\vz{{\bm{z}}}

\def\mI{{\bm{I}}}

\def\mT{{\bm{T}}}

\DeclareMathAlphabet{\mathsfit}{\encodingdefault}{\sfdefault}{m}{sl}
\SetMathAlphabet{\mathsfit}{bold}{\encodingdefault}{\sfdefault}{bx}{n}

\usepackage{hyperref}
\usepackage{url}

\usepackage{booktabs}
\usepackage{multirow}
\usepackage[figurename=Figure]{caption}
\usepackage[export]{adjustbox}

\usepackage{rotating}
\usepackage{listings}

\usepackage{enumitem}

\title{On the Reproducibility of ``FairCLIP: Harnessing Fairness in Vision-Language Learning''}

 \author{\name Hua Chang Bakker\thanks{Equal contribution.} \email  h.c.bakker@uva.nl \\
      University of Amsterdam
      \AND
      \name Angela Madelon Bernardy\footnotemark[1] \email a.m.bernardy@uva.nl \\
      \addr University of Amsterdam
      \AND
      \name Stan Deutekom\footnotemark[1] \email s.c.j.deutekom@uva.nl \\
      University of Amsterdam
      \AND
      \name Stan Fris\footnotemark[1] \email s.c.j.fris@uva.nl \\
      University of Amsterdam}

\def\month{07}  
\def\year{2025} 
\def\openreview{\url{www.your-git-repo.com}} 

\begin{document}

\maketitle

\begin{abstract}
We investigated the reproducibility of FairCLIP, proposed by~\citet{fairclip_luo}, for improving the group fairness of CLIP~\citep{clip-paper} by minimizing image-text similarity score disparities across sensitive groups using the Sinkhorn distance. 
The experimental setup of~\citet{fairclip_luo} was reproduced to primarily investigate the research findings for FairCLIP. 
The model description by~\citet{fairclip_luo} was found to differ from the original implementation. Therefore, a new implementation, A-FairCLIP, is introduced to examine specific design choices.
Furthermore, FairCLIP+ is proposed to extend the FairCLIP objective to include multiple attributes. 
Additionally, the impact of the distance minimization on FairCLIP's fairness and performance was explored.
In alignment with the original authors, CLIP was found to be biased towards certain demographics when applied to zero-shot glaucoma classification using medical scans and clinical notes from the Harvard-FairVLMed dataset.
However, the experimental results on two datasets do not support their claim that FairCLIP improves the performance and fairness of CLIP.
Although the regularization objective reduces Sinkhorn distances, both the official implementation and the aligned implementation, A-FairCLIP, were not found to improve performance nor fairness in zero-shot glaucoma classification.
\end{abstract}

\section{Introduction}
Fairness in machine learning is increasing in importance as the applications of machine learning keep growing.
One of the applications of machine learning is in the medical domain~\citep{GARG2021100370}, where preventing biases is especially important, as this directly impacts patients health. 
Vision-language (VL) models are a type of model increasingly being used in the medical domain~\citep{zhao2024clipmedicalimagingcomprehensive}. 
However, they have also been found to contain biases~\citep{fraser-kiritchenko-2024-examining, lee2023surveysocialbiasvisionlanguage, fairclip_luo}, which introduces the need for the development of unbiased and explainable models.

Several methods have been proposed to reduce biases of machine learning models~\citep{10.1145/3616865, Seth_2023_CVPR}, such as adversarial learning~\citep{berg-etal-2022-prompt, beutel2017datadecisionstheoreticalimplications} and modified loss functions~\citep{NIPS2017_250cf8b5, 10.1007/978-3-642-33486-3_3, prost2019bettertradeoffperformancefairness}.
The latter approach is an in-processing method and can be used while fine-tuning pre-trained VL models for downstream tasks.
For example, \citet{NIPS2017_250cf8b5} use the maximum mean discrepancy~\citep[{MMD;}][]{JMLR:v13:gretton12a} as a regularization term for privileged learning to encourage the distributions of groups to be close.

To mitigate biases in a post-hoc manner, \citet{fairclip_luo} proposed FairCLIP, which uses a regularization function to improve group fairness for CLIP~\citep{clip-paper}, and study the fairness of the model on an extra data set.

In this work, we first provide a detailed outline of the specific claims made by~\citet{fairclip_luo} and reproduce the experiments presented in their original paper. 
We then propose FairCLIP+, a generalized FairCLIP objective designed to improve group fairness for multiple sensitive attributes at the same time.
Additionally, a more detailed analysis of FairCLIP was performed to study the relation between the regularization objective and the performance of the model.
Following this, we extend the evaluation of the FairCLIP method by applying it to a new data set, providing insight into the generalizability.

\section{Scope of reproducibility}
\label{sec:claims}

In order to address biases in CLIP-based models, \citet{fairclip_luo} proposed FairCLIP, a method that uses an optimal transport-based regularization function to improve the fairness of CLIP, particularly in the medical domain. 
This approach aims to minimize the distance between the population distribution and the group distribution of a sensitive attribute.

The focus of our work is on reproducing the following claims made by~\citet{fairclip_luo}:
\begin{enumerate}
    \item CLIP shows significant biases towards 1) Asian, 2) male, 3) non-Hispanic and 4) Spanish-speaking individuals on the Harvard-FairVLMed dataset, but finetuning on the data can alleviate the biases, improving fairness and performance.
    \item Fine-tuning CLIP using the FairCLIP objective on the Harvard-FairVLMed data set improves both performance and fairness of zero-shot glaucoma classification across various subgroups in the Harvard-FairVLMed data set.
\end{enumerate}

\section{Methodology}

\subsection{Model descriptions}
\label{sec:model_descriptions}

\paragraph{CLIP}
CLIP~\citep{clip-paper} is a contrastive vision-language model consisting of a language model and a vision transformer, and can be used to compute a similarity score between an image and a text. 
The similarity scores can be utilized in downstream tasks, such as captioning or classification.
Experiments were carried out on CLIP with the RS50, ViT-B/16 or ViT-L/14 architecture, which have 102M, 150M and 428M parameters, respectively.
All CLIP-based models were initialized using the official checkpoints.

\paragraph{BLIP-2} BLIP-2~\citep{li2023blip2} is another vision-language model, which combines a frozen vision encoder and a frozen language model using a Querying transformer (Q-Former) model.
The Q-former learns the interaction between the vision encoder and the language model.
Following~\citet{fairclip_luo}, a frozen CLIP vision encoder was used for BLIP-2.

\paragraph{FairCLIP} The FairCLIP model is a pre-trained CLIP model that is fine-tuned with the standard CLIP loss and a regularizer for group fairness. 
Group fairness requires that groups are treated equally, which in particular should hold for sensitive attributes.
Let \(\mathbb{A}\) denote the set of sensitive attributes and let \(\mathcal{A} \in \mathbb{A}\) be a sensitive attribute; \(\mathcal{A} = \{\mathrm{male}, \mathrm{female}\}\) for example.
Suppose we have a batch \(\{(\vx_I^{(i)}, \vx_T^{(i)}, a^{(i)})\}^{n}_{i=1}\) containing image features \(\vx^{(i)}_I\), text features \(\vx_T^{(i)}\) and a sensitive group label \(a^{(i)} \in \mathcal{A}\).
Each sensitive group should have the same underlying distribution \(\mathcal{D}_{(\vx_I, \vx_T, a \mid a = \alpha)}\) of similarity scores \(\{ \langle \vx_I^{(i)}, \vx_T^{(i)} \rangle \}^n_{i=1}\), where \(\langle \cdot, \cdot \rangle\) is the Euclidean inner product.
That is, the distribution of diagonal elements of \(\mI \mT^\top\) for an image feature matrix \(\mI\) and text feature matrix \(\mT\).
Computing the actual distributions is intractable; the distributions are thus estimated using a batch \(\mathcal{D}_{\mathcal{B}}\) and a batch \(\mathcal{D}_{\mathcal{B}_a}\) given a sensitive group \(a\).
The closeness of probability distributions can be measured using a distance function \(d\), such as the Sinkhorn distance~\citep{NIPS2013_af21d0c9} or the MMD~\citep{JMLR:v13:gretton12a}.
To improve group fairness, \citet{fairclip_luo} proposed the regularizer
\begin{align}
\label{eq:fairclip_objective}
        \mathcal{L}_{\mathrm{Fair}_\mathcal{A}} = \sum_{\alpha \in \mathcal{A}} d\left( \mathcal{D}_{(\vx_I, \vx_T, a)}, \mathcal{D}_{(\vx_I, \vx_T, a \mid a = \alpha)} \right),
\end{align}
where \(\mathcal{D}_{(\vx_I, \vx_T, a)}\) is the batch distribution.
The FairCLIP regularizer is obtained by setting \(d\) to be the Sinkhorn distance:
\begin{align}
    d(\mathcal{D}_{\mathcal{B}}, \mathcal{D}_{\mathcal{B}})
    =
    \inf_{z } \left[ \mathbb{E}_{(\vp, \vq) \sim z} [c(\vp, \vq)] + \epsilon H (\vz \mid \mathcal{D}_{\mathcal{B}} \otimes \mathcal{D}_{\mathcal{B}_a}) \right],
\end{align}
where \(\vz\) is a joint probability distribution that marginalizes to both \(\mathcal{D}_{\mathcal{B}}\) and \(\mathcal{D}_{\mathcal{B}_a}\), \(c\) is the transport cost, \(\epsilon\) is a regularization parameter, \(H\) is the relative entropy, and \(\mathcal{D}_{\mathcal{B}} \otimes \mathcal{D}_{\mathcal{B}_a}\) is the product measure. 
The regularizer in Equation~\ref{eq:fairclip_objective} is an upper bound on the pairwise distance between the batch distributions of any two groups \(a\in \mathcal{A}\) and \(a' \in \mathcal{A}\).

FairCLIP is fine-tuned using the (FairCLIP) loss
\begin{align}
\label{eq:fairclip_full_loss}
\mathcal{L}_{\mathrm{FairCLIP}_\mathcal{A}} = \mathcal{L}_{\mathrm{CLIP}} + \lambda \mathcal{L}_{\mathrm{Fair}_\mathcal{A}},
\end{align}
where \(\mathcal{L}_{\mathrm{CLIP}}\) is the standard CLIP loss, \(\lambda\) is the regularization rate and \(\mathcal{L}_{\mathrm{Fair}_\mathcal{A}}\) is the FairCLIP regularizer for a sensitive attribute \(\mathcal{A}\).

\paragraph{FairCLIP+} The FairCLIP regularizer only considers a single sensitive attribute during the fine-tuning phase, but it can be beneficial to fine-tune the model to be fair across multiple attributes. 
While it is possible to use the FairCLIP objective with multiple attributes by collapsing sensitive attributes into a single attribute, this may result in a (more) imbalanced data set.
Instead, the FairCLIP regularizer is modified to include multiple attributes using a weighted average and is referred to as the FairCLIP+ regularizer.
More concretely, the weight \(w_{\mathcal{A}}\) of an attribute \(\mathcal{A}\) is restricted to be non-negative and all weights should sum to one.
This restriction on the weights was chosen to limit the effect of including multiple attributes on the size of the regularization value.
The FairCLIP+ regularizer is then defined as
\begin{align}
\label{eq:generalized_fairclip_objective}
        \mathcal{L}_{\mathrm{Fair+}}
        = \sum_{\mathcal{A} \in \mathbb{A}}\sum_{\alpha \in \mathcal{A}} w_{\mathcal{A}} d\left( \mathcal{D}_{(\vx_I, \vx_T, a)}, \mathcal{D}_{(\vx_I, \vx_T, a \mid a = \alpha)} \right),
\end{align}
where \(d\) is the Sinkhorn distance, following~\citet{fairclip_luo}.
This approach is equivalent to adding the FairCLIP regularizer for multiple attributes with separate regularization rates.
Note that setting \(w_{\mathcal{A}} = 0\) ignores the attribute \(\mathcal{A}\), so that the regularizer can be used to fine-tune CLIP on a subset of the sensitive attributes. 
Similar to the FairCLIP loss in Equation~\ref{eq:fairclip_full_loss}, the FairCLIP+ loss is obtained by adding \(\mathcal{L}_{\mathrm{Fair+}}\) to the CLIP loss:
\begin{align}
\label{eq:fairclip_full_loss}
\mathcal{L}_{\mathrm{FairCLIP+}} = \mathcal{L}_{\mathrm{CLIP}} + \lambda \mathcal{L}_{\mathrm{Fair+}}.
\end{align}

\subsection{Data sets}
\label{sec:data set}

\begin{figure}[h]
    \centering
    \includegraphics[width=0.8\textwidth]{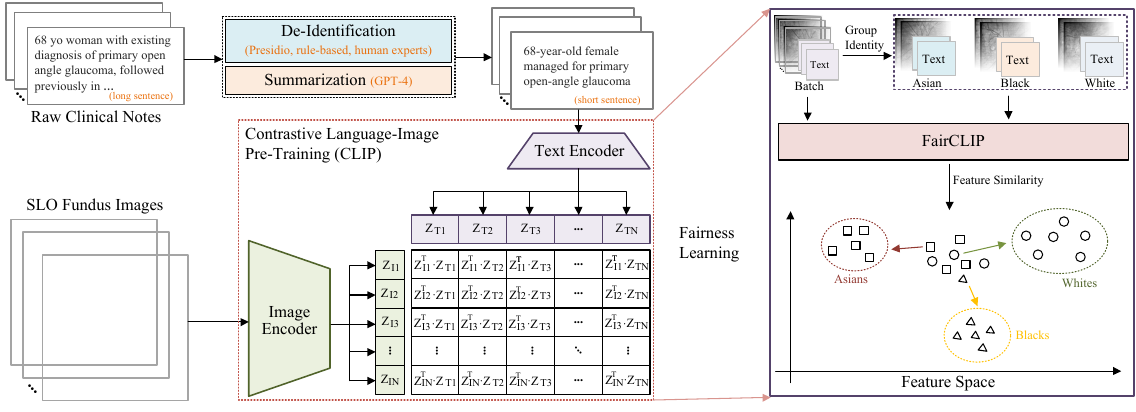}
    \caption{FairCLIP pipeline for fine-tuning on the attribute race~\citep[Figure 2]{fairclip_luo}. A sample is taken for each group to compute the Sinkhorn distances.}
    \label{fig:fairclip_arch}
\end{figure}

\paragraph{Harvard-FairVLMed}
The Harvard-FairVLMed data set\footnote{The data set can be obtained from the GitHub repository of~\citet{fairclip_luo}: \url{https://github.com/Harvard-Ophthalmology-AI-Lab/FairCLIP}.}, collected by~\citet{fairclip_luo}, contains 10,000 samples consisting of 1) scanning laser ophthalmoscopy (SLO) fundus images, 2) clinical notes containing a diagnosis summarized by GPT-4~\citep{openai2024gpt4technicalreport}, and 3) demographic attributes. 
The demographic attributes are i) `gender', ii) `race', iii) `ethnicity', iv) preferred `language', v) `age', and vi) `marital status'.
Following~\citet{fairclip_luo}, we only consider the first four attributes.
Examples of SLO fundus images are shown in Figure~\ref{fig:fairclip_arch}.
It should be noted that the data set is imbalanced for protected attributes. 
For example, the dominant subgroups of the attributes `ethnicity' and preferred `language' make up 95.8\% and 97.3\% of the respective distributions. 
The attributes `race' and `gender' are relatively balanced.
The data set was split into a training, validation and test set containing 7000, 1000 and 2000 samples, respectively. Further analysis of the data set, including the distribution of protected attributes and distributions per split, can be found in Appendix~\ref{appendix:descriptives_data:harvard}. 

\paragraph{FairFace}
\label{subsubsec:data set/fairface}
The generalizability of FairCLIP was investigated using zero-shot prediction experiments on the FairFace data set~\citep{karkkainenfairface}. 
This data set contains 100,000+ manually annotated face images originating from tweets, media photographs and protests, balanced across races and gender.
FairFace is widely used for bias investigation in vision models~\citep{clip-paper, zheng2022general, gustafson2023facet, shahbazi2023representation}. 
The data set is balanced across `race' and `gender' attributes, and thus allows for analysis of fairness. More details on the data set can be found in Appendix~\ref{appendix:descriptives_data:FairFace}.

\subsection{Experimental setup and code}
\label{section:code}
The official implementation for FairCLIP by~\citet{fairclip_luo} is available on Github\footnote{\url{https://github.com/Harvard-Ophthalmology-AI-Lab/FairCLIP}} and was used as a basis for the reproduction of the experiments. 
However, it was found that the implementation of the FairCLIP regularizer differs from its mathematical formulation as described in the original paper.
In particular, the similarity scores of a sensitive group \(\mathcal{D}_{\mathcal{B}_a}\) are divided by their sum, before computing the distance between the batch distribution and the group distribution in the implementation, which is not a part of the model description in~\citet{fairclip_luo}.
Additionally, the official implementation does not use the proposed similarity scores \(\operatorname{diag}(\mI \mT^\top)\) of the text-image pairs, but instead uses the diagonal elements of \(\mI \mT^\top \mT \mI^\top\), where \(\mI\) and \(\mT\) are matrices containing corresponding image and text features, respectively. 
Consequently, the similarity scores in the implementation are more dependent on the used batches.
Finally, model selection was done on the test set rather than the validation set.

We evaluated the method using the official code and an aligned implementation\footnote{All code used in this reproducibility study is available on GitHub: \url{https://github.com/fairclipreproducibility/fairclip-reproducibility}} in which the above deviations are changed to align with the model descriptions. 
Model selection was changed to use the correct dataset to study reproducibility of the findings in both implementations.
Following~\citet{fairclip_luo}, all experiments were repeated three times.

\subsection{Hyperparameters}
\label{section:hyperparameter}
All versions of CLIP and FairCLIP were fine-tuned using contrastive learning, using the same optimizer and loss for fine-tuning on zero-shot prediction as reported by~\citet{fairclip_luo}. 

\paragraph{Parameters for official code} The hyperparameters found by~\citet{fairclip_luo} were used, as our own hyperparameter search produced inferior results on the validation set.
The models were optimized using the Adam optimizer~\citep{kingma2017adammethodstochasticoptimization}, with the parameters set by~\citet{fairclip_luo} in combination with the cross-entropy loss.
The Sinkhorn and MMD distances were computed using the Geomloss library~\citep{feydy2019interpolating} with a batch size of 32 and sample batches of size 32.

The linear probes were trained with a batch size of 512 samples, using learning rate 0.1 and no weight decay, as reported by~\citet{fairclip_luo}.
The linear probes were optimized using the binary cross-entropy loss and the LARS optimizer~\citep{you2017largebatchtrainingconvolutional} for 500 epochs.
Due to resource constraints we used 500 epochs only for the linear probes, while~\citet{fairclip_luo} use 1000 epochs. 
Linear probes with the same setup were also trained on BLIP-2, after fine-tuning for 50 epochs using the hyperparameters reported by~\citep{fairclip_luo}.
The BLIP-2 models were initialized using a checkpoint provided by the Lavis library~\citep{li-etal-2023-lavis}.

\paragraph{Parameters for FairCLIP+} The hyperparameters of the FairCLIP+ model were the same as the FairCLIP parameters reported by \citet{fairclip_luo}, except for the fairness regularizer rate, which was set to $1\cdot 10^{-5}$ rather than $1\cdot 10^{-7}$, as a result of hyperparameter optimization as discussed in Appendix~\ref{appendix:lambda} FairCLIP+ has the additional attribute weight parameters \(\{w_{\mathcal{A}} : \mathcal{A} \in \mathbb{A}\}\).
Due to the imbalances in the dataset, the weights of the attributes `race' and `gender' were set to \(1/2\), and the weights of the remaining attributes were set to \(0\).
This choice of weights is based on strong imbalance of the distributions of attributes `ethnicity' and `language', as shown in Appendix~\ref{appendix:descriptives_data:harvard}.

\paragraph{Parameters for models trained using aligned code} Next to the implementation by \citet{fairclip_luo}, an aligned implementation was created where the implementation was matched with model descriptions, as discussed in section~\ref{section:code}. 
The hyperparameters were optimized using the Optuna library~\citep{optuna_2019}. 
The details of this hyperparameter tuning can be found in Appendix~\ref{appendix:lambda}. 
FairCLIP models fine-tuned on attribute \(\mathcal{A}\) with the aligned implementation are denoted \(\text{A-FairCLIP}_{\mathcal{A}}\).

\paragraph{Alternative distance functions} We have also experimented with replacing the Sinkhorn distance with the Gaussian and Laplacian MMD when fine-tuning the A-FairCLIP models, however inspecting distances showed that this change did not affect the performance of the models across all the parameter ranges used during hyperparameter optimization.
More details can be found in Appendix~\ref{appendix:distance_analysis}.

\subsubsection{Evaluation metrics} 
\label{subsub:eval_metrics}
The fairness of the models was evaluated using the demographic parity distance (DPD) and the difference in equalized odds (DEOdds).
The performance of the models was measured using the area under the receiver operating characteristic curve (AUC), which can also be calculated per group (group-wise AUC), or rescaled per sensitive attribute \(\mathcal{A}\) resulting in the equity-scaled AUC (ES-AUC), which can be used to analyze the fairness performance trade-off.
The ES-AUC~\citep{fairclip_luo} is also a fairness metric and defined by
\begin{align}
    \mathrm{ES-AUC}_{\mathcal{A}} = \frac{\mathrm{AUC}}{1 - \sum_{a \in \mathcal{A}} |\mathrm{AUC} - \mathrm{AUC}_a|},
\end{align}
where \(\mathrm{AUC}_a\) is the group-wise AUC for group \(a \in \mathcal{A}\).

\subsubsection{FairCLIP in the medical domain} 
We followed the procedure described by~\citet{fairclip_luo}.
The CLIP-based models were fine-tuned using a constrastive loss, with an added regularization term for the FairCLIP and FairCLIP+ models, using the GPT-4 summarized notes and unlabelled SLO fundus images per batch.
The FairCLIP models were pre-trained with respect to each sensitive attribute, resulting in four different FairCLIP models. 
The CLIP models fine-tuned without the FairCLIP regularizer are denoted by CLIP-FT.
The fine-tuned models were then used for linear probing and zero-shot classification.

\paragraph{Linear probing} 
The bias of the visual features of CLIP was analyzed by training a linear probe on top of a frozen vision encoder~\citep{10.5555/3045118.3045167}.
Similarly, linear probes for BLIP-2 were trained.
Both CLIP and BLIP-2 use the ViT-L/14 architecture.
The adapted implementation of~\citet{mae_paper} by~\citet{fairclip_luo} was used for training all linear probes.

\paragraph{Zero-shot glaucoma prediction} The transferability and fairness of CLIP-based models in the medical domain were analyzed by performing glaucoma prediction in a zero-shot setting. 
In this experiment, the FairCLIP models were fine-tuned on the sensitive attributes `race', `gender', `ethnicity', and `language', alongside a fine-tuned CLIP model, CLIP-FT, as a baseline.
During zero-shot glaucoma prediction, visual features extracted from SLO fundus images are paired with text embeddings representing possible diagnostic outcomes. 
Specifically, the positive class is represented by the text \verb+A photo of glaucoma+, and the negative class by \verb+A photo of non-glaucoma+. 
The models, trained using contrastive learning, compute similarity scores between image and text embeddings to classify images as either glaucoma-positive or glaucoma-negative, based on the highest similarity score.

\subsubsection{Fairness in CLIP gender prediction}
In order to investigate generalization of the FairCLIP approach, the zero-shot prediction experiments were also performed on the FairFace data set. 
As discussed in Section~\ref{subsubsec:data set/fairface}, this data set contains balanced groups which allows for bias investigation and training. 
Experiments were performed on CLIP, FairCLIP balancing across `race', and FairCLIP+ balancing across both `age' and `race'. 

\paragraph{Fine-tuning}
Fine-tuning was done using contrastive learning in the same way as discussed in~\citet{fairclip_luo}. However, the data set did not contain language descriptions; therefore custom prompts were constructed, as prompting has been shown to improve zero-shot transfer results~\citep{clip-paper}. The prompt is constructed as follows:

\verb+Image of a person that is [AGE] years old, their race is: [RACE] They are [GENDER]+

In this prompt, each variable is replaced with the respective attribute for that instance. This prompting allows for the maximum amount of contrast between each group, with 126 unique possible prompts, distributed equally across the data set. 
Fine-tuning was performed with a random subset of 11,000 samples, this number was chosen to account for memory constraints in used hardware.
For the same reason the RS50 CLIP architecture was used, as this is the smallest available model architecture for CLIP.

\paragraph{Evaluation} 
Evaluation was performed on the full validation set of 10,954 images for all models. Here, a similar prompt was constructed out of the target label: \verb+Image of a [GENDER] person+.
The evaluations was done with all model versions using the same hyperparameters found using the FairVLMed-dataset as discussed in Section~\ref{section:hyperparameter}.
Subsequently, the assigned probabilities per image were used to calculate the metrics discussed in Section~\ref{subsub:eval_metrics}.

\subsection{Computational requirements}
Experiments were carried out on a single NVIDIA A100 40GB GPU along with 9 Intel Xeon Platinum 8360Y CPUs.
For the results on the Harvard-FairVLMed data set, a total of approximately 979 hours  of computation were performed on the A100 GPU, in contrast to approximately 11 hours for the FairFace data set.  
These combined experiments result in an estimated 80.3 kgCO\(_2\)eq emissions. 
The estimate was calculated using the Machine Learning Impact calculator\footnote{\url{https://mlco2.github.io/impact\#compute}} as presented in~\citet{lacoste2019quantifying}. Through Trees For All\footnote{\url{https://treesforall.nl/en/offset-carbon-emissions/}}, 100 kgCO\(_2\)eq were manually offset. 
More details on computational requirements for individual experiments are provided in the Appendix~\ref{appendix:resources}.

\section{Results}
\label{sec:results}
The experimental setup was reproduced using the official code as well as the aligned code, but results showed no conclusive support for the second claim in Section~\ref{sec:claims}.
Linear probing of CLIP showed weak evidence for the claim that CLIP is biased towards certain demographics, and that fine-tuning can reduce these visual biases.
The results for zero-shot classification suggest that the FairCLIP objective effectively reduces the distance between the population and group distributions, but does not necessarily lead to improved fairness or performance. 
The results obtained by~\citet{fairclip_luo} are included in Appendix~\ref{appendix:result_comparison} for easier comparison with the original study.

\subsection{Results reproducing original paper}

\subsubsection{Fairness of the visual features}
\label{sec:finetuning}
Linear probes were trained to verify whether CLIP is visually biased towards certain groups and that fine-tuning on the Harvard-FairVLMed data set can reduce these biases.
Results for CLIP and fine-tuned CLIP (CLIP-FT), as well as BLIP-2 and fine-tuned BLIP-2 (BLIP-2-FT) are shown in Table~\ref{tab:linear_probing_clip:L14}.
With respect to the fairness metrics DPD and DEOdds, CLIP-FT yields better scores than CLIP on the groups `ethnicity' and `language'.
The fine-tuned CLIP model achieves the best AUC and ES-AUC scores, except on the attribute `language', but has a large standard deviation across all attributes.
Both CLIP and CLIP-FT have higher performance on subgroups `Asian', `male' and `non-Hispanic'.

Similar to CLIP, fine-tuning BLIP-2 (BLIP-2-FT) also improves performance AUC scores on the attributes `race', `gender' and `ethnicity'.
BLIP-2-FT performs worse overall on the fairness metrics DPD and DEOdds when compared to BLIP-2. As with CLIP-FT, BLIP-2-FT fails to improve the ES-AUC score for the attribute `language'.
Scores for BLIP-2-based models also show high standard deviations.  

These results align with the claim that CLIP shows bias towards certain groups. 
The results show weak evidence for fine-tuning CLIP improving ES-AUC, due to the high standard deviations and weak results from BLIP-2-FT, which uses the same vision encoder as CLIP.

\begin{table}[t]
\caption{Results of linear probing on CLIP and fine-tuned CLIP models with ViT-L/14 architecture. The highest average scores per model architecture (CLIP or BLIP-2) are marked bold. The models with the -FT suffix are fine-tuned on the Harvard-FairVLMed dataset.
}
\label{tab:linear_probing_clip:L14}
\begin{center}
\adjustbox{width=\textwidth}{
\begin{tabular}{llrrrrrrr}
\toprule
\textbf{Attribute} & \textbf{Model}  & \textbf{DPD $\downarrow$} & \textbf{DEOdds $\downarrow$} & \textbf{AUC $\uparrow$} & \textbf{ES-AUC $\uparrow$} & \multicolumn{3}{c}{\textbf{Group-wise AUC $\uparrow$}} \\ \midrule
&&&&&&\textbf{Asian} & \textbf{Black} & \textbf{White} \\
\multirow{2}{*}{\textbf{Race}} 
 & CLIP & 5.56 $\pm$ 0.76 & \textbf{9.22} $\pm$ 0.94 & 76.48 $\pm$ 0.04 & 71.42 $\pm$ 0.40 & 80.11 $\pm$ 0.02 & 73.38 $\pm$ 0.62 & 76.85 $\pm$ 0.10 \\
 & CLIP-FT & \textbf{2.60} $\pm$ 1.24 & 10.16 $\pm$ 3.79 & \textbf{78.96} $\pm$ 2.87 & \textbf{75.67} $\pm$ 4.03 & \textbf{81.17} $\pm$ 1.42 & \textbf{77.34} $\pm$ 2.60 & \textbf{79.55} $\pm$ 2.90 \\
 \cmidrule{2-9}
& BLIP-2 & {9.59} $\pm$ 0.64 & \textbf{11.13} $\pm$ 1.11 & {73.21} $\pm$ 0.03 & {68.21} $\pm$ 0.28 & {76.31} $\pm$ 0.31 & {69.22} $\pm$ 0.15 & {73.45} $\pm$ 0.02 \\
 & BLIP-2-FT & \textbf{4.04} $\pm$ 1.52 & {11.59} $\pm$ 1.34 & \textbf{78.81} $\pm$ 1.79 & \textbf{71.91} $\pm$ 1.87 & \textbf{82.83} $\pm$ 0.27 & \textbf{74.05} $\pm$ 0.39 & \textbf{79.62} $\pm$ 1.86 \\

                       \midrule
&&&&&&\textbf{Female} & \textbf{Male} \\
\multirow{2}{*}{\textbf{Gender}} 
 & CLIP & \textbf{0.44} $\pm$ 0.02 & \textbf{4.15} $\pm$ 0.26 & 76.48 $\pm$ 0.04 & 71.72 $\pm$ 0.08 & 73.47 $\pm$ 0.03 & 80.11 $\pm$ 0.14 &   \\
 & CLIP-FT & 1.56 $\pm$ 1.57 & 6.54 $\pm$ 1.27 & \textbf{78.96} $\pm$ 2.87 & \textbf{74.83} $\pm$ 2.24 & \textbf{76.44} $\pm$ 2.51 & \textbf{81.95} $\pm$ 3.37 &   \\
 \cmidrule{2-8}
 & BLIP-2 & \textbf{0.23} $\pm$ 0.14 & \textbf{4.26} $\pm$ 0.94 & {73.21} $\pm$ 0.03 & {68.64} $\pm$ 0.07 & {70.18} $\pm$ 0.06 & {76.84} $\pm$ 0.05 &   \\
 & BLIP-2-FT & {2.38} $\pm$ 0.46 & {8.29} $\pm$ 0.51 & \textbf{78.81} $\pm$ 1.79 & \textbf{73.52} $\pm$ 1.57 & \textbf{75.51} $\pm$ 1.68 & \textbf{82.69} $\pm$ 1.83 &   \\

                       \midrule
&&&&&&\textbf{Non-Hispanic} & \textbf{Hispanic} \\
\multirow{2}{*}{\textbf{Ethnicity}} 
 & CLIP & 18.40 $\pm$ 1.67 & 20.35 $\pm$ 1.89 & 76.48 $\pm$ 0.04 & 71.19 $\pm$ 0.34 & 76.69 $\pm$ 0.05 & 69.25 $\pm$ 0.52 &   \\
 & CLIP-FT & \textbf{13.62} $\pm$ 1.96 & \textbf{17.88} $\pm$ 1.07 & \textbf{78.96} $\pm$ 2.87 & \textbf{73.00} $\pm$ 4.60 & \textbf{79.27} $\pm$ 2.77 & \textbf{70.97} $\pm$ 5.77 &   \\
 \cmidrule{2-8}
& BLIP-2 & \textbf{7.76} $\pm$ 2.12 & \textbf{13.66} $\pm$ 3.41 & {73.21} $\pm$ 0.03 & {67.60} $\pm$ 0.11 & {73.54} $\pm$ 0.04 & {65.23} $\pm$ 0.17 &   \\
 & BLIP-2-FT & {14.39} $\pm$ 2.91 & {19.04} $\pm$ 5.07 & \textbf{78.81} $\pm$ 1.79 & \textbf{74.41} $\pm$ 2.88 & \textbf{79.03} $\pm$ 1.68 & \textbf{73.08} $\pm$ 3.41 &   \\

\midrule                        
&&&&&&\textbf{English} & \textbf{Spanish} & \textbf{Others} \\
\multirow{2}{*}{\textbf{Language}} 
 & CLIP & 21.36 $\pm$ 3.89 & 28.53 $\pm$ 8.84 & 76.48 $\pm$ 0.04 & \textbf{71.77} $\pm$ 0.16 & 76.38 $\pm$ 0.02 & \textbf{81.82} $\pm$ 0.00 & \textbf{75.36} $\pm$ 0.20 \\
 & CLIP-FT & \textbf{9.69} $\pm$ 3.55 & \textbf{22.58} $\pm$ 7.01 & \textbf{78.96} $\pm$ 2.87 & 69.12 $\pm$ 1.81 & \textbf{79.45} $\pm$ 3.00 & 79.36 $\pm$ 1.64 & 67.15 $\pm$ 1.74 \\
 \cmidrule{2-9}
& BLIP-2 & {23.63} $\pm$ 0.14 & \textbf{18.77} $\pm$ 0.95 & {73.21} $\pm$ 0.03 & \textbf{70.08} $\pm$ 0.56 & {72.74} $\pm$ 0.04 & {73.77} $\pm$ 0.43 & \textbf{76.65} $\pm$ 0.45 \\
 & BLIP-2-FT & \textbf{12.77} $\pm$ 1.72 & {20.26} $\pm$ 2.26 & \textbf{78.81} $\pm$ 1.79 & {68.48} $\pm$ 1.47 & \textbf{79.32} $\pm$ 1.87 & \textbf{75.95} $\pm$ 3.13 & {68.56} $\pm$ 3.25 \\

\bottomrule
\end{tabular}
}
\end{center}
\end{table}

\subsubsection{Zero-shot glaucoma classification}
\label{sec:zeroshotglaucoma}
The claimed fairness-performance tradeoff, i.e.\ the second claim, was evaluated using a zero-shot glaucoma classification task.
Results are shown in Table~\ref{tab:zero_shot_clip:B16:summarized} for the ViT-B/16 architecture. 
Results for the ViT-L/14 architecture are included in Appendix~\ref{appendix:zero_shot:l14}.
FairCLIP with the ViT-B/16 architecture yields better DPD and DEOdds scores than CLIP-FT on the attributes `ethnicity' and `language', but worse on `race' and `gender'.
Thus, FairCLIP performs better on the fairness metrics of highly imbalanced classes.
When comparing the performance of FairCLIP and CLIP-FT, FairCLIP has worse AUC scores on all attributes.
The extent to which the FairCLIP scores are worse differs per attribute; difference in scores compared to CLIP-FT are often around 3 when considering AUC and ES-AUC.
However, the difference in group-wise AUC for the group `Asian' is 1, whereas the difference for the group `Spanish' is around 12. 
Standard deviations are relatively large.

\begin{table}[t]
\caption{Zero-shot results for fine-tuned CLIP, FairCLIP, and FairCLIP+ models with ViT-B/16 architecture on fairness attributes. The best score is bolded per fairness metric (aligned code is not taken into account as it is not comparable).}
\label{tab:zero_shot_clip:B16:summarized}
\begin{center}
\adjustbox{width=\textwidth}{
\begin{tabular}{llrrrrrrr}
\toprule
\textbf{Attribute} & \textbf{Model}  & \textbf{DPD $\downarrow$} & \textbf{DEOdds $\downarrow$} & \textbf{AUC $\uparrow$} & \textbf{ES-AUC $\uparrow$} & \multicolumn{3}{c}{\textbf{Group-wise AUC $\uparrow$}} \\ \midrule
&&&&&&\textbf{Asian} & \textbf{Black} & \textbf{White} \\
\multirow{3}{*}{\textbf{Race}} 
 & CLIP-FT  & \textbf{8.75} $\pm$ 5.72 & {12.14} $\pm$ 7.02 & \textbf{70.14} $\pm$ 1.63 & \textbf{65.32} $\pm$ 1.55 & \textbf{73.39} $\pm$ 3.01 & {71.48} $\pm$ 1.82 & \textbf{68.73} $\pm$ 2.18 \\
 & FairCLIP\(_{race}\)  & 11.77 $\pm$ 6.82 & 14.89 $\pm$ 3.26 & {66.34} $\pm$ 1.60 & 60.34 $\pm$ 2.23 & {72.39} $\pm$ 1.74 & 68.41 $\pm$ 2.58 & {64.45} $\pm$ 1.65 \\
 & FairCLIP+ & {10.07} $\pm$ 2.56 & \textbf{11.99} $\pm$ 7.51 & {67.78} $\pm$ 2.09 & {61.06} $\pm$ 2.97 & {69.81} $\pm$ 6.95 & \textbf{72.55} $\pm$ 0.91 & {65.91} $\pm$ 2.15 \\
 \cmidrule{2-9} 
 & A-FairCLIP\(_{race}\) & 12.63 $\pm$ 11.03 & 12.77 $\pm$ 11.14 & 59.54 $\pm$ 4.79 & 56.03 $\pm$ 3.37 & 61.31 $\pm$ 5.82 & 61.93 $\pm$ 6.65 & 57.53 $\pm$ 4.69 \\
                       \midrule
&&&&&&\textbf{Female} & \textbf{Male} \\
\multirow{3}{*}{\textbf{Gender}} 
  & CLIP-FT & \textbf{2.97} $\pm$ 3.08 & \textbf{6.24} $\pm$ 6.54 & \textbf{70.14} $\pm$ 1.63 & \textbf{65.62} $\pm$ 0.89 & \textbf{67.17} $\pm$ 1.16 & \textbf{74.05} $\pm$ 2.21 &   \\
 & FairCLIP\(_{gender}\)  & 5.69 $\pm$ 1.23 & 9.43 $\pm$ 1.80 & 65.69 $\pm$ 1.98 & 62.10 $\pm$ 1.34 & 63.21 $\pm$ 1.58 & 69.00 $\pm$ 2.48 &   \\
  & FairCLIP+ & {4.35} $\pm$ 1.01 & {9.01} $\pm$ 1.09 & {67.78} $\pm$ 2.09 & {65.09} $\pm$ 2.36 & {66.25} $\pm$ 2.23 & {70.40} $\pm$ 2.43 &   \\
   \cmidrule{2-9} 
& A-FairCLIP\(_{gender}\) & 1.57 $\pm$ 2.13 & 3.41 $\pm$ 4.36 & 70.00 $\pm$ 1.94 & 66.07 $\pm$ 2.00 & 67.37 $\pm$ 2.04 & 73.31 $\pm$ 1.80 &   \\
                       \midrule
&&&&&&\textbf{Non-Hispanic} & \textbf{Hispanic} \\
\multirow{3}{*}{\textbf{Ethnicity}} 
 & CLIP-FT & 11.36 $\pm$ 4.85 & 16.09 $\pm$ 5.59 & \textbf{70.14} $\pm$ 1.63 & {63.39} $\pm$ 0.83 & \textbf{70.49} $\pm$ 1.70 & {59.84} $\pm$ 1.58 &   \\
 & FairCLIP\(_{ethnicity}\)  & 5.53 $\pm$ 5.50 & 11.41 $\pm$ 7.66 & 65.33 $\pm$ 4.15 & 59.50 $\pm$ 3.09 & 65.69 $\pm$ 4.20 & 55.92 $\pm$ 2.99 &   \\
 & FairCLIP+ & \textbf{2.89} $\pm$ 3.29 & \textbf{10.26} $\pm$ 4.29 & {67.78} $\pm$ 2.09 & \textbf{64.19} $\pm$ 1.72 & {67.64} $\pm$ 1.89 &\textbf{ 73.23} $\pm$ 6.81 &   \\
  \cmidrule{2-9} 
 & A-FairCLIP\_{ethnicity} & 5.41 $\pm$ 3.49 & 8.60 $\pm$ 6.75 & 62.21 $\pm$ 9.53 & 58.86 $\pm$ 6.89 & 62.41 $\pm$ 9.68 & 57.02 $\pm$ 5.68 &   \\
\midrule                        
&&&&&&\textbf{English} & \textbf{Spanish} & \textbf{Others} \\
\multirow{3}{*}{\textbf{Language}} 
 & CLIP-FT & {7.35} $\pm$ 6.05 & 17.39 $\pm$ 11.50 & \textbf{70.14} $\pm$ 1.63 & \textbf{60.64} $\pm$ 1.83 & \textbf{70.30} $\pm$ 1.71 & {62.79} $\pm$ 1.86 & \textbf{61.97} $\pm$ 3.06 \\
 & FairCLIP\(_{language}\)  & \textbf{5.25} $\pm$ 5.04 & \textbf{9.33} $\pm$ 9.15 & 64.23 $\pm$ 1.60 & 52.87 $\pm$ 1.01 & 64.37 $\pm$ 1.69 & 50.85 $\pm$ 1.14 & 56.22 $\pm$ 3.32 \\
 & FairCLIP+ & {20.19} $\pm$ 3.33 & {33.29} $\pm$ 11.14 & {67.78} $\pm$ 2.09 & {55.31} $\pm$ 3.37 & {67.44} $\pm$ 1.90 & \textbf{73.65} $\pm$ 12.06 & {54.67} $\pm$ 4.30 \\
  \cmidrule{2-9} 

  & A-FairCLIP$_{language}$ & 10.16 $\pm$ 6.90 & 14.39 $\pm$ 9.16 & 71.04 $\pm$ 1.99 & 62.24 $\pm$ 0.87 & 71.40 $\pm$ 2.24 & 69.70 $\pm$ 1.46 &  58.86 $\pm$ 2.59 \\

\bottomrule
\end{tabular}
}
\end{center}
\end{table}

\subsection{Results beyond original paper}
To further investigate the support for the claims, we also 1) analyzed design and implementation choices of the FairCLIP regularizer using the aligned code described in Section~\ref{section:code}, 2) considered the effect of taking into account multiple sensitive attributes during the fine-tuning phase with FairCLIP+, 3) investigated whether fine-tuning on attribute \(\mathcal{A}\) affects the fairness or performance on another attribute \(\mathcal{A}'\), 4) analyzed the distances of sensitive groups after fine-tuning and 5) applied the method in a different domain using the FairFace dataset.

\subsubsection{Design choices}

\paragraph{Aligned implementation} The A-FairCLIP models were also evaluated using the aligned code that incorporates the changes described in Section~\ref{section:code}.
Table~\ref{tab:zero_shot_clip:B16:summarized} contains the results for FairCLIP with the ViT-B/16 architecture on the zero-shot glaucoma classification task.
When comparing the A-FairCLIP models with the FairCLIP models, we see that the former type of models have 1) worse performance on all metrics, except DEOdds, for the attribute `race', 2) improved performance on the attribute `gender', which can be explained by the hyperparameter optimization being performed on the attribute gender, 3) improved fairness scores, but worse AUC scores for `ethnicity', and 4) improved AUC scores, but worse fairness scores on the attribute `language'.
On the attribute `gender', we see that the A-FairCLIP models are able to match or beat CLIP-FT on `gender', except for `male' group-wise AUC, as CLIP-FT yields a score of 74.05 and A-FairCLIP 73.31.
The `language' A-FairCLIP models are also able to slightly improve upon CLIP-FT on most AUC-based metrics, but not DPD. 
On the remaining attributes, A-FairCLIP still performs noticeably worse than CLIP-FT.
As with the FairCLIP models, the A-FairCLIP models suffer from high standard deviations. 

\subsubsection{Further fairness analysis}
\label{sec:furtherfairnessanalysis}

\paragraph{FairCLIP+}
Results for FairCLIP+ are shown in Table~\ref{tab:zero_shot_clip:B16:summarized} for the ViT-B/16 architecture. 
The results for the ViT-L/14 architecture can be found in Appendix~\ref{appendix:zero_shot:l14}.
Due to the large standard deviation and small differences in the means of Table~\ref{tab:zero_shot_clip:B16:summarized}, we observe that the FairCLIP+ model trained on `race' and `gender' did not significantly improve scores over the FairCLIP or CLIP-FT models. 
There appear to be small improvements with regards to `ethnicity'; however, these are likely insignificant due to the high standard deviations. 
When only comparing to FairCLIP, it can be seen that FairCLIP+ slightly improves upon FairCLIP across all attributes, although this is likely insignificant due to the high standard deviations as well.
Thus, fine-tuning for multiple fairness attributes simultaneously does not appear to influence the AUC or fairness. 

\paragraph{Performance on non-fine-tuned attributes}
The FairCLIP models in the zero-shot classification experiment were trained with respect to a single sensitive attribute \(\mathcal{A}\).
We investigated whether this also affects the performance of the model on other sensitive attributes.
Table~\ref{tab:zero_shot_clip:B16} shows the performance of FairCLIP models on attributes that were not explicitly taken into account during fine-tuning.
The standard deviations of the scores are rather large; often exceeding the score differences.
This is the case for the DPD and DEOdds scores, for example.
The FairCLIP model fine-tuned on the attribute `language' generally has the best performance on the fairness metrics. 
Surprisingly, models fine-tuned on attribute \(\mathcal{A}\) do not necessarily yield the best performance for \(\mathcal{A}\).
The FairCLIP model fine-tuned for language has the best overall DPD and DEOdds scores. 
When considering the AUC-scores, FairCLIP fine-tuned for `race' achieves the best overall scores. 
The results for the ViT-L/14 architecture can be found in Appendix~\ref{appendix:zero_shot:l14}, and show similar results.

\begin{table}[h]
\caption{Zero-shot results for FairCLIP models fine-tuned on different attributes with ViT-B/16 architecture on fairness attributes. The best scores are bolded per fairness metric.}
\label{tab:zero_shot_clip:B16}
\begin{center}
\adjustbox{width=\textwidth}{
\begin{tabular}{llrrrrrrr}
\toprule
\textbf{Attribute} & \textbf{Model}  & \textbf{DPD $\downarrow$} & \textbf{DEOdds $\downarrow$} & \textbf{AUC $\uparrow$} & \textbf{ES-AUC $\uparrow$} & \multicolumn{3}{c}{\textbf{Group-wise AUC $\uparrow$}} \\ \midrule
&&&&&&\textbf{Asian} & \textbf{Black} & \textbf{White} \\
\multirow{4}{*}{\textbf{Race}} 
 & FairCLIP\(_{race}\) & {11.77} $\pm$ 6.82 & {14.89} $\pm$ 3.26 & \textbf{66.34} $\pm$ 1.60 & 60.34 $\pm$ 2.23 & \textbf{72.39} $\pm$ 1.74 & {68.41} $\pm$ 2.58 & \textbf{64.45} $\pm$ 1.65 \\
 & FairCLIP\(_{gender}\) & 18.61 $\pm$ 2.19 & 18.50 $\pm$ 2.41 & {65.69} $\pm$ 1.98 & 59.27 $\pm$ 2.39 & {70.37} $\pm$ 0.85 & \textbf{69.81} $\pm$ 1.94 & 63.61 $\pm$ 1.94 \\
 & FairCLIP\(_{ethnicity}\) & 13.72 $\pm$ 8.48 & 18.46 $\pm$ 9.17 & 65.33 $\pm$ 4.15 & {60.43} $\pm$ 3.54 & 69.70 $\pm$ 5.11 & 67.61 $\pm$ 4.79 & {63.88} $\pm$ 4.52 \\
 & FairCLIP\(_{language}\) & \textbf{6.77} $\pm$ 9.88 & \textbf{7.04} $\pm$ 9.76 & 64.23 $\pm$ 1.60 & \textbf{60.78} $\pm$ 0.75 & 65.82 $\pm$ 2.64 & 66.72 $\pm$ 2.11 & 62.65 $\pm$ 1.50 \\

                       \midrule
&&&&&&\textbf{Female} & \textbf{Male} \\
\multirow{4}{*}{\textbf{Gender}} 
  & FairCLIP\(_{race}\) & 3.08 $\pm$ 2.31 & 6.16 $\pm$ 2.52 & \textbf{66.34} $\pm$ 1.60 & \textbf{62.65} $\pm$ 1.10 & \textbf{63.78} $\pm$ 1.25 & \textbf{69.67} $\pm$ 2.10 &   \\
 & FairCLIP\(_{gender}\) & 5.69 $\pm$ 1.23 & 9.43 $\pm$ 1.80 & {65.69} $\pm$ 1.98 & {62.10} $\pm$ 1.34 & {63.21} $\pm$ 1.58 & {69.00} $\pm$ 2.48 &   \\
 & FairCLIP\(_{ethnicity}\) & \textbf{1.78} $\pm$ 1.98 & {4.47} $\pm$ 4.62 & 65.33 $\pm$ 4.15 & 61.87 $\pm$ 4.21 & 62.88 $\pm$ 4.36 & 68.50 $\pm$ 3.92 &   \\
 & FairCLIP\(_{language}\) & {1.86} $\pm$ 2.90 & \textbf{3.53} $\pm$ 4.04 & 64.23 $\pm$ 1.60 & 60.67 $\pm$ 1.82 & 61.68 $\pm$ 1.81 & 67.54 $\pm$ 1.42 &   \\

                       \midrule
&&&&&&\textbf{Non-Hispanic} & \textbf{Hispanic} \\
\multirow{4}{*}{\textbf{Ethnicity}} 
 & FairCLIP\(_{race}\) & {5.38} $\pm$ 3.49 & {8.54} $\pm$ 3.43 & \textbf{66.34} $\pm$ 1.60 & {60.17} $\pm$ 0.57 & \textbf{66.71} $\pm$ 1.74 & {56.43} $\pm$ 1.96 &   \\
 & FairCLIP\(_{gender}\) & 9.53 $\pm$ 4.24 & 14.24 $\pm$ 0.91 & {65.69} $\pm$ 1.98 & \textbf{60.39} $\pm$ 2.18 & {66.02} $\pm$ 1.93 & \textbf{57.22} $\pm$ 2.64 &   \\
 & FairCLIP\(_{ethnicity}\) & 5.53 $\pm$ 5.50 & 11.41 $\pm$ 7.66 & 65.33 $\pm$ 4.15 & 59.50 $\pm$ 3.09 & 65.69 $\pm$ 4.20 & 55.92 $\pm$ 2.99 &   \\
 & FairCLIP\(_{language}\) & \textbf{4.93} $\pm$ 6.61 & \textbf{5.88} $\pm$ 8.07 & 64.23 $\pm$ 1.60 & 58.14 $\pm$ 1.80 & 64.61 $\pm$ 1.64 & 54.11 $\pm$ 2.57 &   \\

\midrule                        
&&&&&&\textbf{English} & \textbf{Spanish} & \textbf{Others} \\
\multirow{4}{*}{\textbf{Language}} 
 & FairCLIP\(_{race}\) & 11.75 $\pm$ 0.62 & 17.26 $\pm$ 4.23 & \textbf{66.34} $\pm$ 1.60 & \textbf{57.36} $\pm$ 1.16 & \textbf{66.46} $\pm$ 1.72 & {57.39} $\pm$ 3.55 & \textbf{59.73} $\pm$ 1.61 \\
 & FairCLIP\(_{gender}\) & 15.99 $\pm$ 1.35 & 29.04 $\pm$ 0.65 & {65.69} $\pm$ 1.98 & {55.73} $\pm$ 6.94 & {65.78} $\pm$ 1.82 & 53.03 $\pm$ 11.78 & {59.67} $\pm$ 3.51 \\
 & FairCLIP\(_{ethnicity}\) & {8.22} $\pm$ 5.01 & {17.08} $\pm$ 7.73 & 65.33 $\pm$ 4.15 & 53.86 $\pm$ 4.20 & 65.51 $\pm$ 4.27 & \textbf{60.42} $\pm$ 15.23 & 54.12 $\pm$ 2.04 \\
 & FairCLIP\(_{language}\) & \textbf{5.25} $\pm$ 5.04 & \textbf{9.33} $\pm$ 9.15 & 64.23 $\pm$ 1.60 & 52.87 $\pm$ 1.01 & 64.37 $\pm$ 1.69 & 50.85 $\pm$ 1.14 & 56.22 $\pm$ 3.32 \\

\bottomrule
\end{tabular}
}
\end{center}
\end{table}

\paragraph{Analysis of the Sinkhorn distance}\label{sec:analysis_sinkhorn}
The Sinkhorn distances between subgroups and the population distribution on the test set are shown in Table~\ref{tab:distances_b16}.
Note that both FairCLIP+ as well as both FairCLIP models minimize the distances; their values are much smaller when compared to CLIP or CLIP-FT. 
When fine-tuned on one `attribute', the distance for all groups tends to be significantly smaller when compared to CLIP, which may be due to the groups generally not being disjoint. 
For example, FairCLIP trained on `race', shows small distances for gender groups as well. 
Additionally, FairCLIP\(_{language}\), which scores higher on the fairness metrics for `race' and `gender', does not minimize the distances for these groups the best.
Next to this, CLIP-FT increases the differences in distances for groups of the attributes `race' and `gender'. For example, the distance of subgroup `Black' was half that of the subgroup `Asian' in CLIP, whereas it is almost three times larger in CLIP-FT. 
This indicates that fine-tuning without a fairness metric might even increase biases, which is not in line with the results described in Section~\ref{sec:zeroshotglaucoma}.
Finally, we observe that FairCLIP+, on average, minimizes the distances more than both FairCLIP\(_{race}\) and FairCLIP\(_{language}\), even for attributes that were not a part of its loss function. 
This might correlate with the first finding: if fine-tuning on one attribute decreases the distance of all attributes, fine-tuning on two might increase this effect further. 
Fine-tuning on two attributes also increases the number of samples, for which distances are minimized. 
Furthermore, FairCLIP+ has a larger fairness rate value, $10^{-5}$ instead of $10^{-7}$, enforcing the distance to be further minimized. 
It should also be noted that there are occasionally high standard deviations. 

Overall, these results, in combination with Table~\ref{tab:zero_shot_clip:B16} and Table~\ref{tab:zero_shot_clip:B16:summarized}, indicate that minimizing the Sinkhorn distance does not imply fairer or improved glaucoma prediction when using the Harvard-FairVLMed data set.

\paragraph{Distances of A-FairCLIP}
Distances of the A-FairCLIP model fine-tuned on `race' are shown in Table~\ref{tab:distances_b16_refactored_smaller}.
We see the distances are decreased with regards to CLIP. 
The distances are larger compared to CLIP-FT, except for the subgroup `Asian`.
Again, we do not see a correspondence between smaller distances and better fairness or performance.

The results for MMD distances, as well as a complete overview of the Sinkhorn distances for the aligned code, can be found in Appendix~\ref{appendix:distance_analysis}, where the Sinkhorn distances show similar results, and the MMD distances show no change in distances.

\begin{table}[t]
\caption{Comparison of Sinkhorn distances per group for CLIP, CLIP-FT and A-FairCLIP$\_{race}$ using the ViT-B/16 architecture. Note that CLIP has only been run once, since it has pre-defined weights.}
\label{tab:distances_b16_refactored_smaller}
\begin{center}
\adjustbox{width=0.8\textwidth}{
\begin{tabular}{llrrrr}
\toprule
\textbf{Attribute} & \textbf{Group} & \textbf{CLIP} & \textbf{CLIP-FT} & \textbf{A-FairCLIP\(_{race}\)}\\ 
\midrule
\multirow{3}{*}{\textbf{Race}} 
    & Asian        & $5.04\cdot 10^{-1}$ &$ 2.16\cdot 10^{-1} $$\pm$ $1.37\cdot 10^{-1}$ & $ 6.44 \cdot 10^{-2}$ $\pm $$ 6.34 \cdot 10^{-2}$\\
    & Black        & $3.62\cdot 10^{-1}$ &$ 8.63\cdot 10^{-2} $$\pm$ $5.60\cdot 10^{-2}$ &  $ 1.37 \cdot 10^{-1}$ $\pm $$ 1.18 \cdot 10^{-1}$\\
    & White        & $4.66\cdot 10^{-2}$   &$ 5.82\cdot 10^{-3} $$\pm$ $3.20\cdot 10^{-3} $ & $ 6.95 \cdot 10^{-3}$ $\pm $$ 5.67 \cdot 10^{-3}$\\ 
\bottomrule
\end{tabular}
}
\end{center}
\end{table}

\begin{table}[h]
\caption{Comparison of Sinkhorn distances per group for different CLIP versions using the ViT-B/16 architecture. Note that CLIP has only been run once, since it has pre-defined weights.}
\label{tab:distances_b16}
\begin{center}
\adjustbox{width=\textwidth}{
\begin{tabular}{llrrrrrr}
\toprule
\textbf{Attribute} & \textbf{Group} & \textbf{CLIP} & \textbf{CLIP-FT} & \textbf{FairCLIP+} & \textbf{FairCLIP\(_{race}\)} & \textbf{FairCLIP\(_{language}\)} \\ 
\midrule
\multirow{3}{*}{\textbf{Race}} 
    & Asian        & 3103.06 & 4782.07 $\pm$ 2393.31 & 12.61 $\pm$ 10.88 & 29.23 $\pm$ 32.60 & 30.44 $\pm$ 14.86\\
    & Black        & 1502.70 & 12182.48 $\pm$ 7687.03 & 10.44 $\pm$ 4.96 & 31.07 $\pm$ 36.05 & 49.61 $\pm$ 21.36\\
    & White        & 69.28   & 4.40 $\pm$ 1.94 & 0.07 $\pm$ 0.03 & 0.14 $\pm$ 0.08 & 0.19 $\pm$ 0.04 \\ 
\midrule
\multirow{2}{*}{\textbf{Gender}} 
    & Female       & 706.46  & 27.23 $\pm$ 14.68 & 0.19 $\pm$ 0.03 & 0.84 $\pm$ 0.78 & 0.66 $\pm$ 0.56\\
    & Male         & 117.71  & 563.38 $\pm$ 444.46 & 1.42 $\pm$ 0.71 & 5.42 $\pm$ 3.73 & 11.38 $\pm$ 4.00\\ 
\midrule
\multirow{2}{*}{\textbf{Ethnicity}} 
    & Non-Hispanic & 8.48    & 14.61 $\pm$ 15.55 & 0.22 $\pm$ 0.31 & 0.021 $\pm$ 0.01 & 0.16 $\pm$ 0.19\\
    & Hispanic     & 34478.90 & 2457.55 $\pm$ 258.91 & 10.80 $\pm$ 3.24 & 63.99 $\pm$ 79.89 & 96.21 $\pm$ 91.85\\ 
\midrule
\multirow{3}{*}{\textbf{Language}} 
    & English      & 5.12    & 15.21 $\pm$ 20.23 & 0.01 $\pm$ 0.003 & 0.02 $\pm$ 0.02 & 0.30 $\pm$ 0.41 \\
    & Spanish      & 63284.91 & 28871.51 $\pm$ 17163.83 & 42.26 $\pm$ 10.09 & 442.59 $\pm$ 594.74 & 264.18 $\pm$ 158.70 \\
    & Other        & 20176.54 & 6991.91 $\pm$ 2176.67 & 40.97 $\pm$ 14.48 & 99.34 $\pm$ 119.34 & 78.16 $\pm$ 47.06 \\ 
\bottomrule
\end{tabular}
}
\end{center}
\end{table}

\begin{table}[t]
\caption{Combined zero-shot gender prediction performance of fine-tuned CLIP, FairCLIP+ fine-tuned on `age' and `race', and FairCLIP+ fine-tuned on `race' models. The best score is bolded per fairness metric (per row per attribute).}
\label{tab:fairface_combined}
\begin{center}
\adjustbox{width=\textwidth}{
\begin{tabular}{lrccc|rccc}
\toprule
&       & \multicolumn{3}{c|}{\textbf{Performance by Race Groups}}      &      & \multicolumn{3}{c}{\textbf{Performance by Age Groups}}       \\
                            && \textbf{CLIP-FT}          & \textbf{FairCLIP+\(_{age, race}\)}    & \textbf{FairCLIP\(_{race}\)}      &                           & \textbf{CLIP-FT}          & \textbf{FairCLIP+\(_{age,race}\)}    & \textbf{FairCLIP\(_{race}\)}      \\
\midrule
\parbox[t]{4mm}{\multirow{4}{*}{\rotatebox[origin=c]{90}{\textbf{Metric}}}} 
&\textbf{DPD $\downarrow$}    & 22.71 $\pm$ 0.49          & \textbf{22.29} $\pm$ 1.21  & 22.39 $\pm$ 0.87 &                       & 29.66 $\pm$ 1.29 & \textbf{26.74} $\pm$ 0.94  & 27.03 $\pm$ 2.57          \\
&\textbf{DEOdds $\downarrow$} & 14.12 $\pm$ 1.48          & \textbf{13.05} $\pm$ 2.51  & 13.12 $\pm$ 2.45             &                       & 33.73 $\pm$ 1.66 & 30.31 $\pm$ 5.43           & \textbf{29.71} $\pm$ 5.04 \\
&\textbf{AUC $\uparrow$}      & 98.12 $\pm$ 0.04          & 98.15 $\pm$ 0.02           & \textbf{98.17} $\pm$ 0.04    &                       & 98.12 $\pm$ 0.04 & 98.15 $\pm$ 0.02           & \textbf{98.17} $\pm$ 0.04 \\
&\textbf{ES-AUC $\uparrow$}   & 92.22 $\pm$ 0.17          & 92.39 $\pm$ 0.12           & \textbf{92.66} $\pm$ 0.31    &                       & 77.92 $\pm$ 0.28 & 77.85 $\pm$ 0.39           & \textbf{78.02} $\pm$ 0.47 \\
\midrule
&\textbf{Race Group}          &                           &                            &                              & \textbf{Age Group}    &                            &                  &                           \\
\parbox[t]{4mm}{\multirow{9}{*}{\rotatebox[origin=c]{90}{\textbf{Group-wise AUC}}}}
&\textbf{East Asian}          & 98.44 $\pm$ 0.07          & \textbf{98.46} $\pm$ 0.11  & 98.41 $\pm$ 0.04             & \textbf{0-2}          & \textbf{88.50} $\pm$ 0.12  & 88.44 $\pm$ 0.37 & 88.07 $\pm$ 0.70          \\
&\textbf{Indian}              & 98.20 $\pm$ 0.05          & 98.29 $\pm$ 0.05           & \textbf{98.37} $\pm$ 0.11    & \textbf{3-9}          & 91.53 $\pm$ 0.12           & 91.68 $\pm$ 0.17 & \textbf{91.74} $\pm$ 0.32 \\
&\textbf{Black}               & 94.97 $\pm$ 0.18          & 95.07 $\pm$ 0.13           & \textbf{95.26} $\pm$ 0.22    & \textbf{10-19}        & 94.19 $\pm$ 0.16           & 94.28 $\pm$ 0.19 & \textbf{94.36} $\pm$ 0.09 \\
&\textbf{White}               & \textbf{98.81} $\pm$ 0.02 & 98.79 $\pm$ 0.01           & 98.79 $\pm$ 0.03             & \textbf{20-29}        & \textbf{99.07} $\pm$ 0.01  & 99.05 $\pm$ 0.03 & 99.06 $\pm$ 0.04          \\
&\textbf{Middle Eastern}      & 99.34 $\pm$ 0.03          & \textbf{99.36} $\pm$ 0.03  & 99.29 $\pm$ 0.04             & \textbf{30-39}        & \textbf{99.48} $\pm$ 0.02  & 99.47 $\pm$ 0.01 & 99.46 $\pm$ 0.01          \\
&\textbf{Southeast Asian}     & 97.71 $\pm$ 0.03          & 97.79 $\pm$ 0.11           & \textbf{97.79} $\pm$ 0.08    & \textbf{40-49}        & \textbf{99.47} $\pm$ 0.04  & 99.42 $\pm$ 0.01 & 99.45 $\pm$ 0.03          \\
&\textbf{Latino Hispanic}     & 98.63 $\pm$ 0.06          & 98.65 $\pm$ 0.02           & \textbf{98.65} $\pm$ 0.01    & \textbf{50-59}        & 99.18 $\pm$ 0.03           & 99.19 $\pm$ 0.04 & \textbf{99.21} $\pm$ 0.05 \\
&                             &                           &                            &                              & \textbf{60-69}        & \textbf{98.11} $\pm$ 0.04  & 98.00 $\pm$ 0.06 & 98.01 $\pm$ 0.03          \\
&                             &                           &                            &                              & \textbf{70+}          & 97.09 $\pm$ 0.23           & 96.81 $\pm$ 0.23 & \textbf{97.34} $\pm$ 0.18 \\
\bottomrule
\end{tabular}
}
\end{center}
\end{table}

\subsubsection{Generalizability to other data sets}
\label{subsub:generalizabilityresults}
The results of the zero-shot gender prediction experiments on the FairFace data set are shown in Table~\ref{tab:fairface_combined}.
FairCLIP+ was optimized for both attributes 'age' and 'race' (FairCLIP+) and for `race' only (FairCLIP\(_{race}\)). 

The models perform well in general when considering the protected attribute `race': the AUC is above 98\% for all models. 
The difference in group-wise AUC performance on specific race subgroups is the largest for the subgroup `Black' (around 95\%) and `White' (99.8\%).
The FairCLIP+ model fine-tuned on `race' achieves the highest ES-AUC score and slightly improves the imbalanced performance of group-wise AUC scores over the other models.
The FairCLIP+ model that was fine-tuned on all categories, has the second best performance, but has the fairest predictions with respect to DPD and DEOdds.
However, the standard deviations of the models are rather high relative to the differences in performance, which is most noticeable for the DPD and DEOdds metrics.

The fairness and performance of the models with respect to the age groups are shown in Table~\ref{tab:fairface_combined}.
Compared to the results in performance by race groups in Table~\ref{tab:fairface_combined}, there are large differences in performance across groups. 
This is most noticeable in age groups 0-2 and 3-9. 
The models that were not fine-tuned to balance across age groups (CLIP-FT and FairCLIP\(_{race}\)) perform similar or better across all categories compared to the FairCLIP+ model that is trained to improve group fairness for both the attributes `age' and `race'. 
This is seen both performance and fairness metrics on the groups that generally show weak performance  (0-2 and 3-9). There are also high standard deviations compared to the differences in performance across all models. 
These results indicate that the FairCLIP+ objective might not offer improvements in performance or fairness when fine-tuned on the FairFace dataset.

\section{Discussion}

\subsection{Claims}
\subsubsection*{Claim 1: CLIP shows significant biases towards 1) Asian, 2) male, 3) non-Hispanic and 4) Spanish-speaking individuals on the Harvard-FairVLMed dataset, but finetuning on the data can alleviate the biases, improving fairness and performance.}

\paragraph{Biases} In Section~\ref{sec:finetuning}, linear probing experiments with the frozen vision encoder of CLIP showed significant biases towards the same four groups as claimed.
The fine-tuned CLIP models were shown to have improved AUC, ES-AUC and group-wise AUC scores on the attributes `race', `gender' and `ethnicity'. 
However, the fairness metrics DPD and DEOdds did not necessarily improve on the more balanced attributes. 
Linear probing of BLIP-2-FT, which use a frozen vision encoder of CLIP, showed similar results as fairness and performance were not always better than the (non-fine-tuned) BLIP-2.

\paragraph{Support for claim} It should be taken into consideration that 1) standard deviations are relatively high and 2) the attributes `ethnicity' and `language' are highly imbalanced. 
Therefore, the results (weakly) support Claim 1.
In particular, the high standard deviation suggests instability of fine-tuning, as performance has a large dependency on the seed. 
Standard deviations may be stabilized by increasing the number of runs for each experiment.
It should also be noted that~\citet{fairclip_luo} trained the linear probes for 1000 epochs, which may partially explain the higher standard deviations.

\subsubsection*{Claim 2: Fine-tuning CLIP using the FairCLIP objective on the Harvard-FairVLMed data set improves both performance and fairness of zero-shot glaucoma classification across various subgroups in the Harvard-FairVLMed data set.}

\paragraph{Reproduction} 
When using the official code, FairCLIP performed worse than CLIP-FT across all attributes on most metrics. 
Furthermore, it was observed that FairCLIP models fine-tuned for some attribute \(\mathcal{A}\) did not necessarily yield the best performance or fairness score for attribute \(\mathcal{A}\).
However, the standard deviation of the scores was relatively high compared to the difference in scores.
Even though the scores were worse overall, the distance analysis showed that the FairCLIP models were able to minimize the distances effectively using the best hyperparameters found by~\citet{fairclip_luo}.
High standard deviations were also observed during the distance analysis.
Similar observations were made for FairCLIP+, which was fine-tuned on the two most balanced classes: the model had worse fairness and AUC-based scores for the zero-shot classification task compared to CLIP-FT, while having very small group distances.
FairCLIP+ showed small improvements on the attributes that the model was fine-tuned on compared to FairCLIP, but this is unlikely to be significant due to the high standard deviations and may be explained by to the attributes not being fully disjoint, as shown in Appendix~\ref{appendix:descriptives_data:harvard}. 

The implementation of FairCLIP suggested that the selection of the best model in the training loop was performed on the test set, rather than the validation set.
For our experiments, we changed this to the correct data set, but this may help explain why the parameters reported by~\citet{fairclip_luo} outperformed the best parameters during our hyperparameter optimization phase for FairCLIP+ 

\paragraph{Implementation and design choices} The implementation of FairCLIP, i.e.\ A-FairCLIP, was also aligned with the model descriptions to analyze design choices.
Results showed that scores improved for some attributes compared to FairCLIP, but still do not improve significantly over CLIP-FT on most attributes.
Replacing the Sinkhorn distance with the MMD distance in the A-FairCLIP regularizer did not seem to have a significant effect on the distances as discussed in Appendix~\ref{appendix:distance_analysis}. The Sinkhorn distances of A-FairCLIP were minimized compared to CLIP, but this did not show a correspondence between better fairness.

\paragraph{Another dataset} Replicability was also studied using a dataset from another domain.
On the FairFace dataset, no clear performance increase was observed with FairCLIP+.
There are slight performance increases when fine-tuning on 1) `age' and `race' or 2) `race' only, indicating possible improvements in balancing performance and fairness over groups. 
However, as with the other experiments, the standard deviations weaken the support for performance improvements.
The FairCLIP experiments on the Harvard-FairVLMed data set showed that models fine-tuned for a specific attribute do not yield the best performance on the same attribute.
This was also observed on the FairFace data set: the FairCLIP+ model that was fine-tuned with `age' and 'race' performance on the age attribute compared to models that were not fine-tuned on age.

\paragraph{Support for claim} Our experimental results therefore do not support Claim 2, fine-tuning with the FairCLIP or FairCLIP+ objectives did not show improved performance nor fairness compared to CLIP-FT across attributes and datasets.
In particular, the results suggest that minimizing the (Sinkhorn) distance of the subgroups and the population is not necessarily positively related to the performance nor fairness of the FairCLIP or FairCLIP+ models.

\subsection{Communication with original authors}
We reached out to the original authors by email and provided them with preliminary results, as well as asking about modelling choices, but received no response. 

\newpage
\bibliography{main}
\bibliographystyle{tmlr}

\appendix

\newpage
\section{Dataset statistics}
\subsection{Harvard-FairVLMed}
\label{appendix:descriptives_data:harvard}
\begin{figure}[h]
\begin{center}
\includegraphics[width=0.95\textwidth]{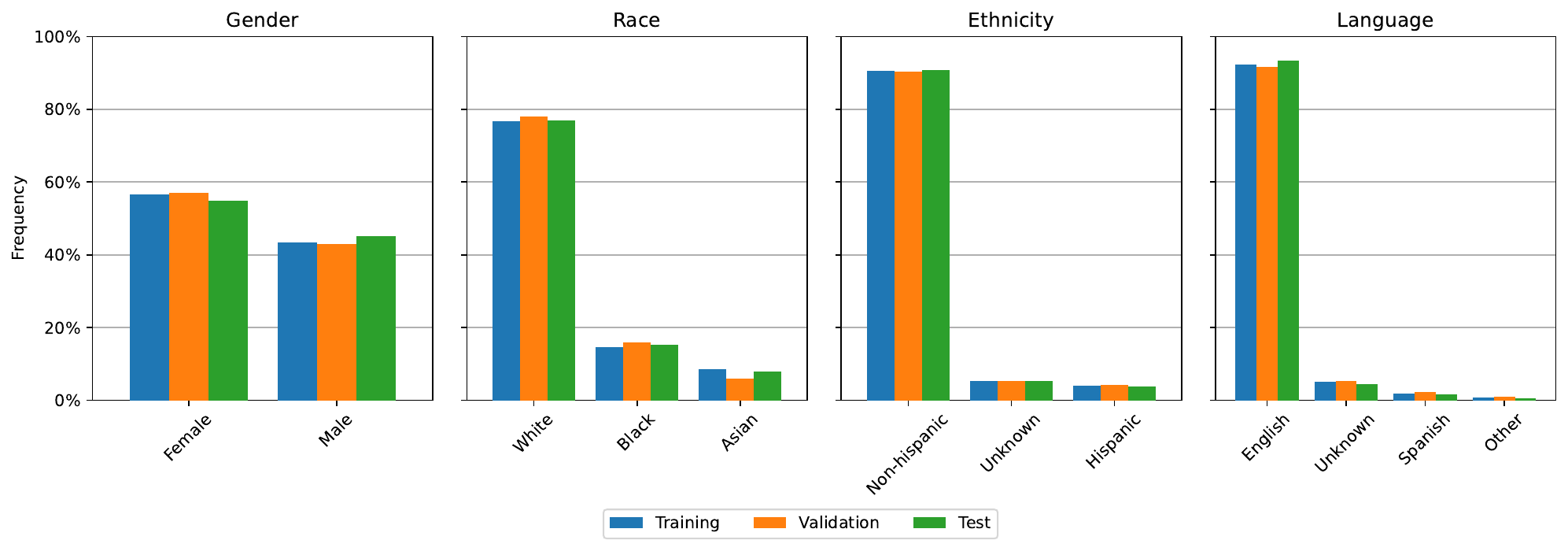}
\end{center}
\caption{Distribution of the sensitive attributes in the Harvard-FairVLMed data set across train, validation and test sets.}
\label{fig:attribute_distribution_all_sets}
\end{figure}

\begin{figure}[h]
\begin{center}
\includegraphics[width=0.95\textwidth]{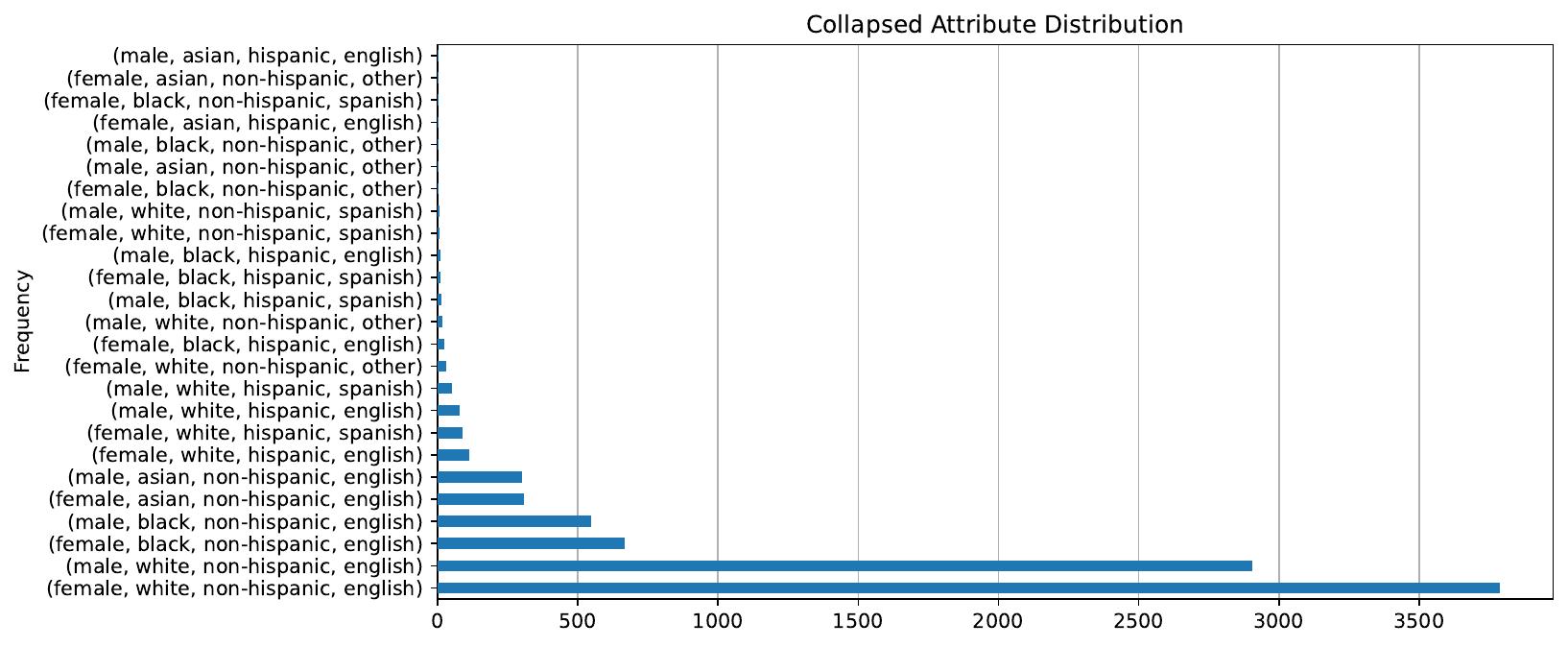}
\end{center}
\caption{Distribution of the collapsed sensitive attributes in the Harvard-FairVLMed data set.}
\label{fig:collapsed_attribute_distribution_non_log}
\end{figure}

The distribution of the data set per sensitive attribute is shown in Figure~\ref{fig:attribute_distribution_all_sets}.
When collapsing the sensitive attributes, i.e.\ combining sensitive groups, the imbalance of the data set increases, as shown in Figure~\ref{fig:collapsed_attribute_distribution_non_log}.
As can be seen, the distribution when considering `ethnicity' and `language' is highly imbalanced.
The subgroup of non-Hispanic English-speakers compromises 94.8\% of the data set.
The two largest groups, consisting of male or female, white non-Hispanic English-speakers, form 74.5\% of the entire data set. 
Important for the FairCLIP+ objective is that distributions between groups are independent from one another, for example in each race group a comparable distribution of gender is preferred, as otherwise the loss between components is not independent. As can be seen in Figure~\ref{fig:collapsed_attribute_distribution_non_log}, the general distribution of attributes does appear to be independent from one another.

\newpage
\subsection{FairFace}
\label{appendix:descriptives_data:FairFace}
\begin{figure}[h]
\begin{center}
\includegraphics[width=0.95\textwidth]{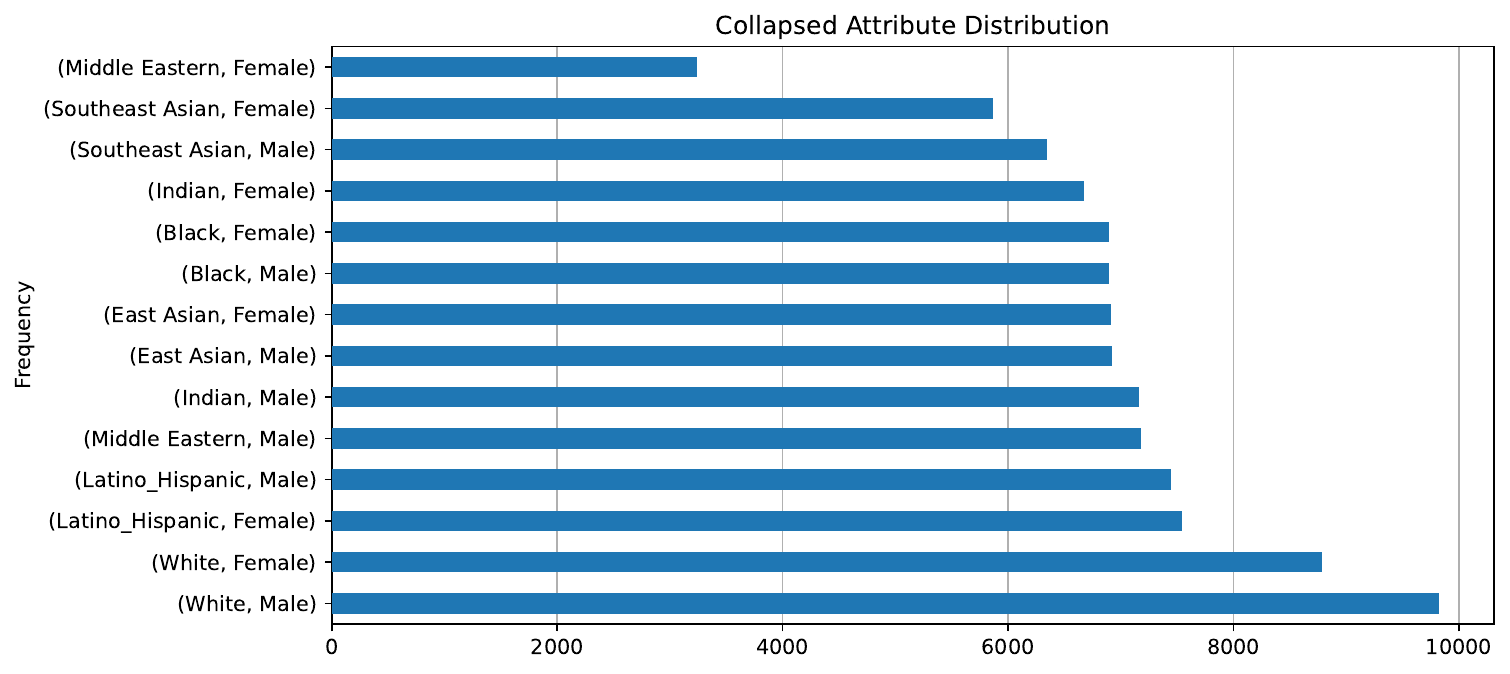}
\end{center}
\caption{Distribution of the collapsed sensitive attributes in the FairFace data set.}
\label{fig:collapsed_attribute_distribution_fairface_non_log}
\end{figure}

The distribution of the collapsed attributes is shown in Figure~\ref{fig:collapsed_attribute_distribution_fairface_non_log}.
Compared to the Harvard-FairVLMed data set, the FairFace data is more balanced across groups.

\section{Hyperparameters}\label{appendix:lambda}
The hyperparameter tuning process involved a grid search over a predefined range of values for the hyperparameters. 
Specifically, the learning rate was selected from the range $[1 \cdot 10^{-8}, 1 \cdot 10^{-2}]$ and the fairness loss rate was chosen from the ranges $[1 \cdot 10^{-7}, 1],\ [0.5, 10],\ [100, 1000]$, with one range per optimization run. The FairCLIP objective was chosen from the set $\{\text{Sinkhorn}, \text{Gaussian MMD}, \text{Laplacian MMD}\}$, and the attribute was chosen from the set $\{\text{gender}, \text{ethnicity}\}$. All parameters for the aligned models used the parameters found with the attribute gender, since there were no significant differences between those parameters and the parameters found with gender. Furthermore, the gender attribute is more balanced, and thus might be more generalizable to models finetuned on other attributes. 
The hyperparameter optimization resulted in three types of aligned models for the Sinkhorn distance. 
The final model has the smallest fairness regularizer rate of $0.07$ and a learning rate of $1.49 \cdot 10^{-5}$, and is denoted as A-FairCLIP$_{\mathcal{A}}$.

The optimization used Optuna's trial mechanism, where each trial sampled a combination of hyperparameter values using a categorical sampling strategy. The objective function evaluated the performance of the model on the validation set, measured by the AUC. The optimizations were carried out for 24 hours running 41 to 150 trials, based on model complexity, each corresponding to a unique hyperparameter configuration.
The best trial achieved a maximum AUC of $0.7283$ with the following hyperparameter configuration: the learning rate was set to $1.49 \cdot 10^{-5}$, the fairness loss weight was set to $0.07$, and the FairCLIP objective was set to the Sinkhorn objective.

\begin{figure}[h]
\begin{center}
\includegraphics[width=\textwidth]{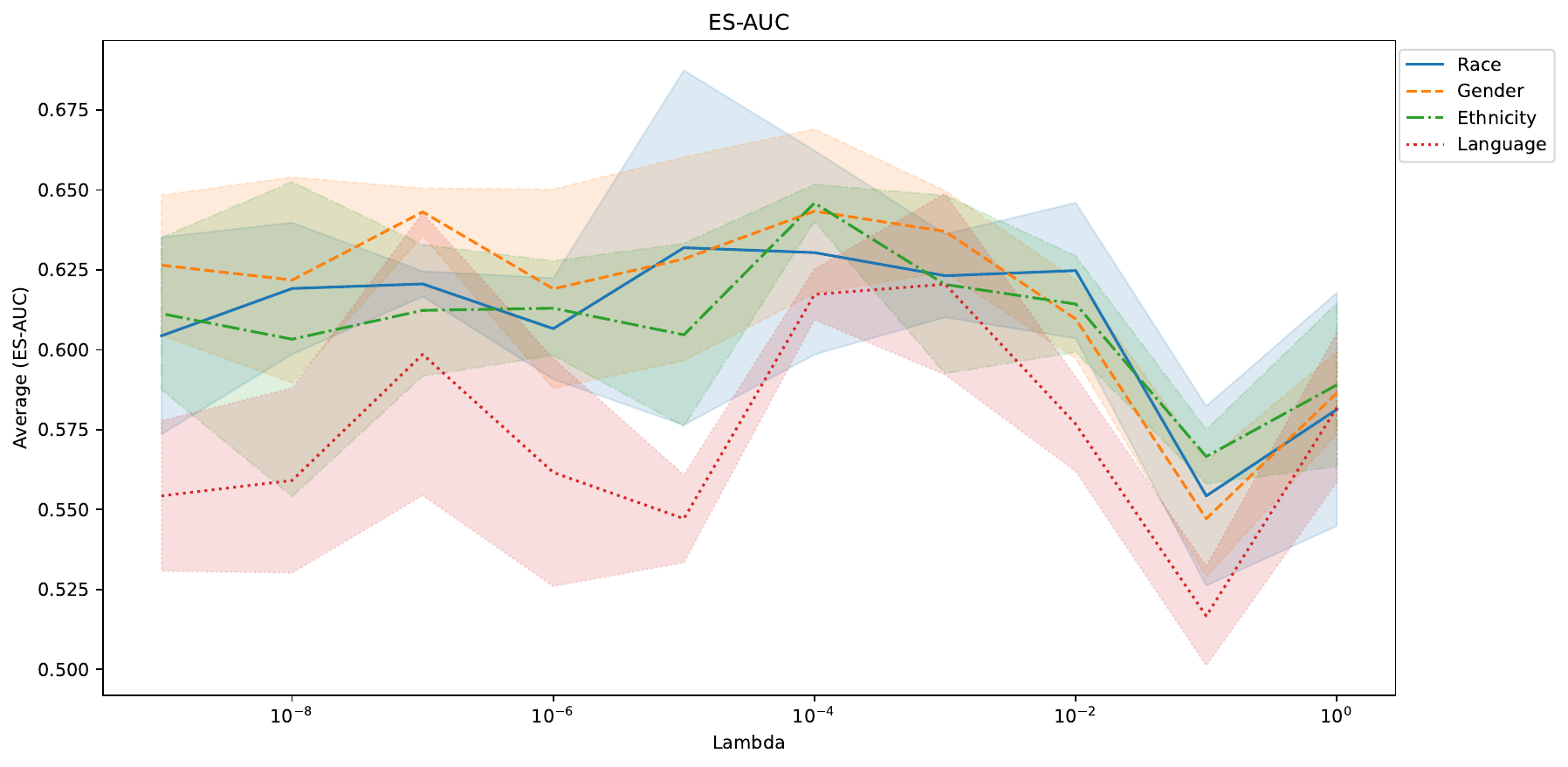}
\end{center}
\caption{ES-AUC scores on FairCLIP\(_{race}\) for different regularization rates $\lambda$ on the validation set and the ViT-B/16 architecture using the original implementation presented by \citet{fairclip_luo}.}
\label{fig:esauc_plots}
\end{figure}

The ES-AUC scores for various regularization rates on the validation set are shown in Figure~\ref{fig:esauc_plots} for FairCLIP fine-tuned on the attribute `race'.
The other parameters are as reported in Section~\ref{section:hyperparameter}.

\newpage
\section{Additional reproducibility results for ViT-L/14 on zero-shot classification task}
\label{appendix:zero_shot:l14}

Zero-shot results are included in Table~\ref{tab:zero_shot_clip:L14}.
When using the ViT-L/14 architecture, FairCLIP achieves a better DPD score for `ethnicity' and `language', better DEOdds scores for `gender', `ethnicity' and `language', and improved group-wise AUC scores for `ethnicity' and `language' compared to CLIP-FT. 
However, it does not improve the fairness nor AUC on the `race' attribute. 
Furthermore, the standard deviations often are larger than the differences between CLIP-FT and FairCLIP.

The results of the FairCLIP variants with ViT-L/14 evaluated on all attributes are included in Table~\ref{tab:zero_shot_clip:L14:cross_attribute_comparison}.
The `language'-variant achieves better scores than the other FairCLIP variants, but not necessarily on the attribute `language'.

\begin{table}[h]
\caption{Zero-shot results for fine-tuned CLIP, FairCLIP and FairCLIP+ models with ViT-L/14 architecture on fairness attributes. The best score is bolded per fairness metric. 
}
\label{tab:zero_shot_clip:L14}
\begin{center}
\adjustbox{width=\textwidth}{
\begin{tabular}{llrrrrrrr}
\toprule
\textbf{Attribute} & \textbf{Model}  & \textbf{DPD $\downarrow$} & \textbf{DEOdds $\downarrow$} & \textbf{AUC $\uparrow$} & \textbf{ES-AUC $\uparrow$} & \multicolumn{3}{c}{\textbf{Group-wise AUC $\uparrow$}} \\ \midrule
&&&&&&\textbf{Asian} & \textbf{Black} & \textbf{White} \\
\multirow{3}{*}{\textbf{Race}} 
 & CLIP-FT  & \textbf{17.26} $\pm$ 2.07 & \textbf{18.25} $\pm$ 4.69 & 66.96 $\pm$ 2.17 & \textbf{64.46} $\pm$ 3.20 & 68.70 $\pm$ 2.16 & 67.41 $\pm$ 0.55 & \textbf{66.06} $\pm$ 2.54 \\
 & FairCLIP\(_{race}\) & 21.88 $\pm$ 6.11 & 25.07 $\pm$ 8.90 & 64.71 $\pm$ 3.41 & 61.05 $\pm$ 2.16 & 67.01 $\pm$ 4.03 & 66.72 $\pm$ 4.59 & 63.06 $\pm$ 3.26 \\
 & FairCLIP+ & {17.80} $\pm$ 2.98 & {19.32} $\pm$ 4.17 & \textbf{67.50} $\pm$ 3.61 & {60.54} $\pm$ 7.14 & \textbf{69.19} $\pm$ 2.33 & \textbf{73.35} $\pm$ 1.56 & {65.21} $\pm$ 4.81 \\

                       \midrule
&&&&&&\textbf{Female} & \textbf{Male} \\
\multirow{3}{*}{\textbf{Gender}} 
 & CLIP-FT & \textbf{4.17} $\pm$ 1.99 & 8.93 $\pm$ 2.33 & 66.96 $\pm$ 2.17 & 63.08 $\pm$ 1.69 & 64.25 $\pm$ 1.80 & 70.40 $\pm$ 2.74 &   \\
 & FairCLIP\(_{gender}\) & 6.14 $\pm$ 1.61 & \textbf{8.66} $\pm$ 1.61 & 64.88 $\pm$ 1.05 & 61.87 $\pm$ 0.95 & 62.78 $\pm$ 1.10 & 67.64 $\pm$ 1.24 &   \\
 & FairCLIP+ & {6.26} $\pm$ 1.91 & 14.00 $\pm$ 2.28 & \textbf{67.50} $\pm$ 3.61 & \textbf{63.18} $\pm$ 4.06 & \textbf{64.81} $\pm$ 4.04 & \textbf{71.71} $\pm$ 2.88 &   \\

                       \midrule
&&&&&&\textbf{Non-Hispanic} & \textbf{Hispanic} \\
\multirow{3}{*}{\textbf{Ethnicity}} 
 & CLIP-FT  & 10.56 $\pm$ 3.23 & 16.64 $\pm$ 2.36 & 66.96 $\pm$ 2.17 & 62.16 $\pm$ 1.14 & 67.22 $\pm$ 2.22 & 59.50 $\pm$ 0.71 &   \\
 & FairCLIP\(_{ethnicity}\)  & 8.96 $\pm$ 5.01 & \textbf{12.59} $\pm$ 10.94 & 67.13 $\pm$ 4.15 & {62.25} $\pm$ 3.09 & 67.41 $\pm$ 4.25 & {59.57} $\pm$ 3.60 &   \\
 & FairCLIP+ & \textbf{6.83} $\pm$ 4.52 & {14.06} $\pm$ 1.27 & \textbf{67.50} $\pm$ 3.61 & \textbf{64.51} $\pm$ 4.64 & \textbf{67.55} $\pm$ 3.39 & \textbf{67.57} $\pm$ 9.54 &   \\

\midrule                        
&&&&&&\textbf{English} & \textbf{Spanish} & \textbf{Others} \\
\multirow{3}{*}{\textbf{Language}} 
 & CLIP-FT & 14.21 $\pm$ 1.45 & 18.76 $\pm$ 3.73 & 66.96 $\pm$ 2.17 & {60.37} $\pm$ 4.01 & 66.92 $\pm$ 2.11 & 63.92 $\pm$ 7.00 & 60.51 $\pm$ 3.67 \\
 & FairCLIP\(_{language}\) & \textbf{8.32} $\pm$ 5.71 & \textbf{16.53} $\pm$ 10.97 & \textbf{68.53} $\pm$ 3.24 & 60.02 $\pm$ 4.14 & \textbf{68.69} $\pm$ 3.29 & 63.54 $\pm$ 6.59 & 59.37 $\pm$ 2.49 \\
 & FairCLIP+ & {21.79} $\pm$ 4.79 & {24.28} $\pm$ 7.56 & {67.50} $\pm$ 3.61 & \textbf{60.76} $\pm$ 5.18 & {67.04} $\pm$ 3.84 & \textbf{73.01} $\pm$ 0.55 & \textbf{62.17} $\pm$ 4.26 \\
\bottomrule
\end{tabular}
}
\end{center}
\end{table}

\begin{table}[h]
\caption{Zero-shot results for FairCLIP models with ViT-L/14 architecture fine-tuned on different fairness attributes. The best score per fairness metric is bolded. 
}
\label{tab:zero_shot_clip:L14:cross_attribute_comparison}
\begin{center}
\adjustbox{width=\textwidth}{
\begin{tabular}{llrrrrrrr}
\toprule
\textbf{Attribute} & \textbf{Model}  & \textbf{DPD $\downarrow$} & \textbf{DEOdds $\downarrow$} & \textbf{AUC $\uparrow$} & \textbf{ES-AUC $\uparrow$} & \multicolumn{3}{c}{\textbf{Group-wise AUC $\uparrow$}} \\ \midrule
&&&&&&\textbf{Asian} & \textbf{Black} & \textbf{White} \\
\multirow{4}{*}{\textbf{Race}} 
 & FairCLIP\(_{race}\) & 21.88 $\pm$ 6.11 & 25.07 $\pm$ 8.90 & 64.71 $\pm$ 3.41 & 61.05 $\pm$ 2.16 & 67.01 $\pm$ 4.03 & 66.72 $\pm$ 4.59 & 63.06 $\pm$ 3.26 \\
 & FairCLIP\(_{gender}\) & 18.46 $\pm$ 5.54 & {19.34} $\pm$ 8.35 & 64.88 $\pm$ 1.05 & 62.63 $\pm$ 0.94 & 65.66 $\pm$ 0.17 & 66.14 $\pm$ 2.49 & 63.99 $\pm$ 0.73 \\
 & FairCLIP\(_{ethnicity}\) & {18.23} $\pm$ 3.76 & 19.97 $\pm$ 8.90 & {67.13} $\pm$ 4.15 & {64.44} $\pm$ 4.39 & {69.00} $\pm$ 3.70 & {67.80} $\pm$ 4.76 & {66.34} $\pm$ 4.21 \\
 & FairCLIP\(_{language}\) & \textbf{9.89} $\pm$ 4.67 & \textbf{9.48} $\pm$ 4.03 & \textbf{68.53} $\pm$ 3.24 & \textbf{65.55} $\pm$ 2.82 & \textbf{71.19} $\pm$ 4.26 & \textbf{68.62} $\pm$ 4.66 & \textbf{67.78} $\pm$ 3.15 \\

                       \midrule
&&&&&&\textbf{Female} & \textbf{Male} \\
\multirow{4}{*}{\textbf{Gender}} 
 & FairCLIP\(_{race}\) & 5.96 $\pm$ 1.76 & 9.30 $\pm$ 4.05 & 64.71 $\pm$ 3.41 & 60.75 $\pm$ 2.82 & 61.95 $\pm$ 3.10 & 68.45 $\pm$ 3.80 &   \\
 & FairCLIP\(_{gender}\) & 6.14 $\pm$ 1.61 & 8.66 $\pm$ 1.61 & 64.88 $\pm$ 1.05 & 61.87 $\pm$ 0.95 & 62.78 $\pm$ 1.10 & 67.64 $\pm$ 1.24 &   \\
 & FairCLIP\(_{ethnicity}\) & {4.57} $\pm$ 2.12 & {6.49} $\pm$ 1.42 & {67.13} $\pm$ 4.15 & {64.04} $\pm$ 3.29 & {65.04} $\pm$ 3.60 & {69.82} $\pm$ 4.72 &   \\
 & FairCLIP\(_{language}\) & \textbf{4.35} $\pm$ 1.79 & \textbf{5.71} $\pm$ 2.11 & \textbf{68.53} $\pm$ 3.24 & \textbf{65.13} $\pm$ 2.16 & \textbf{66.29} $\pm$ 2.53 & \textbf{71.48} $\pm$ 4.06 &   \\

                       \midrule
&&&&&&\textbf{Non-Hispanic} & \textbf{Hispanic} \\
\multirow{4}{*}{\textbf{Ethnicity}} 
 & FairCLIP\(_{race}\) & \textbf{4.74} $\pm$ 5.48 & \textbf{8.29} $\pm$ 3.17 & 64.71 $\pm$ 3.41 & 60.01 $\pm$ 2.18 & 65.01 $\pm$ 3.48 & 57.21 $\pm$ 1.79 &   \\
 & FairCLIP\(_{gender}\) & 6.20 $\pm$ 5.71 & 10.43 $\pm$ 6.55 & 64.88 $\pm$ 1.05 & 59.75 $\pm$ 1.25 & 65.21 $\pm$ 1.04 & 56.61 $\pm$ 1.73 &   \\
 & FairCLIP\(_{ethnicity}\) & 8.96 $\pm$ 5.01 & 12.59 $\pm$ 10.94 & {67.13} $\pm$ 4.15 & {62.25} $\pm$ 3.09 & {67.41} $\pm$ 4.25 & \textbf{59.57} $\pm$ 3.60 &   \\
 & FairCLIP\(_{language}\) & {5.03} $\pm$ 3.76 & {9.61} $\pm$ 5.59 & \textbf{68.53} $\pm$ 3.24 & \textbf{62.44} $\pm$ 0.46 & \textbf{68.86} $\pm$ 3.46 & {59.09} $\pm$ 2.35 &   \\

\midrule                        
&&&&&&\textbf{English} & \textbf{Spanish} & \textbf{Others} \\
\multirow{4}{*}{\textbf{Language}} 
 & FairCLIP\(_{race}\) & 13.40 $\pm$ 7.14 & {16.28} $\pm$ 3.36 & 64.71 $\pm$ 3.41 & \textbf{60.40} $\pm$ 3.50 & 64.45 $\pm$ 3.58 & \textbf{65.62} $\pm$ 0.49 & \textbf{61.52} $\pm$ 2.14 \\
 & FairCLIP\(_{gender}\) & 13.93 $\pm$ 3.26 & 17.56 $\pm$ 10.68 & 64.88 $\pm$ 1.05 & 56.93 $\pm$ 5.87 & 64.87 $\pm$ 0.99 & 54.64 $\pm$ 8.43 & {61.22} $\pm$ 4.61 \\
 & FairCLIP\(_{ethnicity}\) & {12.09} $\pm$ 2.19 & \textbf{12.88} $\pm$ 1.11 & {67.13} $\pm$ 4.15 & 58.13 $\pm$ 3.68 & {67.26} $\pm$ 4.08 & 58.24 $\pm$ 4.13 & 60.66 $\pm$ 4.25 \\
 & FairCLIP\(_{language}\) & \textbf{8.32} $\pm$ 5.71 & 16.53 $\pm$ 10.97 & \textbf{68.53} $\pm$ 3.24 & {60.02} $\pm$ 4.14 & \textbf{68.69} $\pm$ 3.29 & {63.54} $\pm$ 6.59 & 59.37 $\pm$ 2.49 \\

\bottomrule
\end{tabular}
}
\end{center}
\end{table}

\section{Comparison of results with original paper}\label{appendix:result_comparison}
Table~\ref{tab:linear_probing_comparison:L14}, Table~\ref{tab:comparison_original:B16:} and Table~\ref{tab:comparison_original:L14} show the results reported by~\citet{fairclip_luo} and our experimental results, for linear probing and zero-shot tasks.
The results of our linear probing experiments are comparable to the results of~\citet{fairclip_luo}, but have higher standard deviations.
Table~\ref{tab:comparison_original:B16:} and Table~\ref{tab:comparison_original:L14} show similar standard deviations as reported by the original authors.
On the ViT-B/16 architecture, the relationship between CLIP-FT and FairCLIP seems to be inverted in comparison to the results of~\citet{fairclip_luo}. In line with the original paper, we see that FairCLIP ViT-B/16 performs better with respect to CLIP-FT ViT-B/16, whereas FairCLIP ViT-L/14 performs worse with respect to CLIP-FT ViT-L/14. We expect this is due to the hyperparameter optimization being performed on the ViT-B/16 architecture.

\begin{table}[h]
\caption{Results of linear probing on CLIP and fine-tuned CLIP models with ViT-L/14 architecture, compared with the results by \citet[Table 2]{fairclip_luo} (denoted with *).}
\label{tab:linear_probing_comparison:L14}
\begin{center}
\adjustbox{width=\textwidth}{
\begin{tabular}{llrrrrrrr}
\toprule
\textbf{Attribute} & \textbf{Model}  & \textbf{DPD $\downarrow$} & \textbf{DEOdds $\downarrow$} & \textbf{AUC $\uparrow$} & \textbf{ES-AUC $\uparrow$} & \multicolumn{3}{c}{\textbf{Group-wise AUC $\uparrow$}} \\ \midrule
&&&&&&Asian & Black & White \\
\multirow{4}{*}{Race} 
& CLIP* & 5.30 ± 0.63 & 14.00 ± 1.01 & 77.27 ± 0.03 & 72.43 ± 0.29 & 79.74 ± 0.31 & 73.60 ± 0.12 & 77.82 ± 0.03 \\
 & CLIP & 5.56 ± 0.76 & 9.22 ± 0.94 & 76.48 ± 0.04 & 71.42 ± 0.40 & 80.11 ± 0.02 & 73.38 ± 0.62 & 76.85 ± 0.10 \\
& CLIP-FT* & 4.01 ± 0.47 & 9.57 ± 0.83 & 80.27 ± 0.08 & 74.70 ± 0.33 & 82.19 ± 0.26 & 75.67 ± 0.21 & 81.20 ± 0.08 \\
 & CLIP-FT & 2.60 ± 1.24 & 10.16 ± 3.79 & 78.96 ± 2.87 & 75.67 ± 4.03 & 81.17 ± 1.42 & 77.34 ± 2.60 & 79.55 ± 2.90 \\

\midrule
&&&&&&Female & Male \\
\multirow{4}{*}{Gender} 
& CLIP* & 1.08 ± 0.19 & 5.19 ± 0.54 & 77.27 ± 0.03 & 72.47 ± 0.10 & 74.25 ± 0.07 & 80.88 ± 0.03 &  \\
 & CLIP & 0.44 ± 0.02 & 4.15 ± 0.26 & 76.48 ± 0.04 & 71.72 ± 0.08 & 73.47 ± 0.03 & 80.11 ± 0.14 &   \\
& CLIP-FT* & 0.39 ± 0.26 & 4.55 ± 0.33 & 80.27 ± 0.08 & 75.81 ± 0.12 & 77.59 ± 0.09 & 83.47 ± 0.04 &  \\
 & CLIP-FT & 1.56 ± 1.57 & 6.54 ± 1.27 & 78.96 ± 2.87 & 74.83 ± 2.24 & 76.44 ± 2.51 & 81.95 ± 3.37 &   \\

\midrule
&&&&&&Non-Hispanic & Hispanic \\
\multirow{4}{*}{Ethnicity} 
& CLIP* & 15.83 ± 0.42 & 14.73 ± 0.54 & 77.27 ± 0.03 & 71.70 ± 0.08 & 77.51 ± 0.03 & 69.73 ± 0.13 & \\
 & CLIP & 18.40 ± 1.67 & 20.35 ± 1.89 & 76.48 ± 0.04 & 71.19 ± 0.34 & 76.69 ± 0.05 & 69.25 ± 0.52 &   \\
& CLIP-FT* & 14.50 ± 0.72	& 22.49 ± 1.44 & 80.27 ± 0.08 &	76.31 ± 0.36 & 80.48 ± 0.07 & 75.30 ± 0.46 &  \\
 & CLIP-FT & 13.62 ± 1.96 & 17.88 ± 1.07 & 78.96 ± 2.87 & 73.00 ± 4.60 & 79.27 ± 2.77 & 70.97 ± 5.77 &   \\

\midrule                        
&&&&&&English & Spanish & Others \\
\multirow{4}{*}{Language} 
 & CLIP* & 13.57 ± 1.35 & 33.11 ± 0.53 & 77.27 ± 0.03 & 70.89 ± 0.21 & 77.25 ± 0.03 & 84.00 ± 0.16 & 75.02 ± 0.28 \\
 & CLIP & 21.36 ± 3.89 & 28.53 ± 8.84 & 76.48 ± 0.04 & 71.77 ± 0.16 & 76.38 ± 0.02 & 81.82 ± 0.00 & 75.36 ± 0.20 \\
 & CLIP-FT* & 16.75 ± 1.08 & 15.74 ± 0.28 & 80.27 ± 0.08 & 67.06 ± 0.46 & 80.77 ± 0.07 & 74.43 ± 0.99 & 66.91 ± 0.17 \\
 & CLIP-FT & 9.69 ± 3.55 & 22.58 ± 7.01 & 78.96 ± 2.87 & 69.12 ± 1.81 & 79.45 ± 3.00 & 79.36 ± 1.64 & 67.15 ± 1.74 \\

\bottomrule
\end{tabular}
}
\end{center}
\end{table}

\begin{table}[h]
\caption{Zero-shot results for fine-tuned CLIP and FairCLIP models with ViT-B/16 architecture on fairness attributes, compared with the results by \citet[Table 3]{fairclip_luo} (denoted with *).}
\label{tab:comparison_original:B16:}
\begin{center}
\adjustbox{width=\textwidth}{
\begin{tabular}{llrrrrrrr}
\toprule
\textbf{Attribute} & \textbf{Model}  & \textbf{DPD $\downarrow$} & \textbf{DEOdds $\downarrow$} & \textbf{AUC $\uparrow$} & \textbf{ES-AUC $\uparrow$} & \multicolumn{3}{c}{\textbf{Group-wise AUC $\uparrow$}} \\ \midrule
&&&&&&\textbf{Asian} & \textbf{Black} & \textbf{White} \\
\multirow{4}{*}{\textbf{Race}} 
& CLIP-FT* & 15.35 $\pm$ 6.50 & 15.11 $\pm$ 5.01 & 67.84 $\pm$ 0.90 & 61.67 $\pm$ 0.63 & 73.11 $\pm$ 2.74 & 70.78 $\pm$ 1.84 & 66.02 $\pm$ 0.60 \\
 & CLIP-FT  & {8.75} $\pm$ 5.72 & {12.14} $\pm$ 7.02 & 70.14 $\pm$ 1.63 & 65.32 $\pm$ 1.55 & 73.39 $\pm$ 3.01 & {71.48} $\pm$ 1.82 & 68.73 $\pm$ 2.18 \\
& FairCLIP\(_{race}\)* & {6.07} $\pm$ 2.44 & {10.50} $\pm$ 2.73 & {70.24} $\pm$ 1.26 & 65.50 $\pm$ 2.60 & {74.83} $\pm$ 0.46 & 71.39 $\pm$ 0.66 & {69.17} $\pm$ 2.10 \\
 & FairCLIP\(_{race}\)  & 11.77 $\pm$ 6.82 & 14.89 $\pm$ 3.26 & {66.34} $\pm$ 1.60 & 60.34 $\pm$ 2.23 & {72.39} $\pm$ 1.74 & 68.41 $\pm$ 2.58 & {64.45} $\pm$ 1.65 \\

                       \midrule
&&&&&&{Female} & {Male} \\
\multirow{4}{*}{{Gender}} 
& CLIP-FT* & 4.34 $\pm$ 0.66 & 9.95 $\pm$ 0.64 & 67.84 $\pm$ 0.90 & 63.21 $\pm$ 0.83 & 64.62 $\pm$ 0.83 & 71.96 $\pm$ 1.22
 \\
  & CLIP-FT & 2.97 $\pm$ 3.08 & {6.24} $\pm$ 6.54 & 70.14 $\pm$ 1.63 & 65.62 $\pm$ 0.89 & 67.17 $\pm$ 1.16 & 74.05 $\pm$ 2.21 &   \\
& FairCLIP\(_{gender}\)* & {0.84} $\pm$ 0.25 & {2.97} $\pm$ 2.07 & {69.76} $\pm$ 2.49 & 65.39 $\pm$ 2.39 & 66.81 $\pm$ 2.50 & {73.50} $\pm$ 2.41 \\
 & FairCLIP\(_{gender}\) & 5.69 $\pm$ 1.23 & 9.43 $\pm$ 1.80 & 65.69 $\pm$ 1.98 & 62.10 $\pm$ 1.34 & 63.21 $\pm$ 1.58 & 69.00 $\pm$ 2.48 &   \\

                       \midrule
&&&&&&\textbf{Non-Hispanic} & \textbf{Hispanic} \\
\multirow{4}{*}{\textbf{Ethnicity}} 
& CLIP-FT* & 8.86 $\pm$ 5.95 & 15.33 $\pm$ 5.18 & 67.84 $\pm$ 0.90 & 63.09 $\pm$ 0.26 & 68.08 $\pm$ 0.97 & 60.56 $\pm$ 0.30 \\
 & CLIP-FT & 11.36 $\pm$ 4.85 & 16.09 $\pm$ 5.59 & {70.14} $\pm$ 1.63 & {63.39} $\pm$ 0.83 & 70.49 $\pm$ 1.70 & {59.84} $\pm$ 1.58 &   \\
& FairCLIP\(_{ethnicity}\)* & 9.12 $\pm$ 3.25 & 14.30 $\pm$ 7.20 & 69.30 $\pm$ 1.87 & 63.33 $\pm$ 0.25 & 69.62 $\pm$ 2.02 & 60.19 $\pm$ 1.18 \\
 & FairCLIP\(_{ethnicity}\) & 5.53 $\pm$ 5.50 & 11.41 $\pm$ 7.66 & 65.33 $\pm$ 4.15 & 59.50 $\pm$ 3.09 & 65.69 $\pm$ 4.20 & 55.92 $\pm$ 2.99 &   \\

\midrule                        
&&&&&&\textbf{English} & \textbf{Spanish} & \textbf{Others} \\
\multirow{4}{*}{\textbf{Language}} 
& CLIP-FT* & 11.11 $\pm$ 2.42 & 16.97 $\pm$ 2.73 & 67.84 $\pm$ 0.90 & 60.19 $\pm$ 3.47 & 67.88 $\pm$ 1.15 & 61.93 $\pm$ 4.57 & 61.89 $\pm$ 2.90 \\
 & CLIP-FT & {7.35} $\pm$ 6.05 & 17.39 $\pm$ 11.50 & 70.14 $\pm$ 1.63 & 60.64 $\pm$ 1.83 & 70.30 $\pm$ 1.71 & {62.79} $\pm$ 1.86 & 61.97 $\pm$ 3.06 \\
& FairCLIP\(_{language}\)* & {7.34} $\pm$ 4.63 & 17.15 $\pm$ 11.13 & {70.08} $\pm$ 1.14 & 62.31 $\pm$ 0.96 & {70.22} $\pm$ 1.27 & 68.47 $\pm$ 5.49 & 63.47 $\pm$ 2.13 \\
 & FairCLIP\(_{language}\) & 5.25 $\pm$ 5.04 & {9.33} $\pm$ 9.15 & 64.23 $\pm$ 1.60 & 52.87 $\pm$ 1.01 & 64.37 $\pm$ 1.69 & 50.85 $\pm$ 1.14 & 56.22 $\pm$ 3.32 \\

\bottomrule
\end{tabular}
}
\end{center}
\end{table}

\begin{table}[h]
\caption{Zero-shot results for fine-tuned CLIP and FairCLIP models with ViT-L/14 architecture on fairness attributes, compared with the results by \citet[Table 3]{fairclip_luo} (denoted with *).
}
\label{tab:comparison_original:L14}
\begin{center}
\adjustbox{width=\textwidth}{
\begin{tabular}{llrrrrrrr}
\toprule
\textbf{Attribute} & \textbf{Model}  & \textbf{DPD $\downarrow$} & \textbf{DEOdds $\downarrow$} & \textbf{AUC $\uparrow$} & \textbf{ES-AUC $\uparrow$} & \multicolumn{3}{c}{\textbf{Group-wise AUC $\uparrow$}} \\ \midrule
&&&&&&\textbf{Asian} & \textbf{Black} & \textbf{White} \\
\multirow{4}{*}{\textbf{Race}} 
 & CLIP-FT* & 10.10 $\pm$ 9.44 & 10.79 $\pm$ 10.41 & 67.83 $\pm$ 2.92 & 63.53 $\pm$ 1.83 & 70.65 $\pm$ 4.58 & 70.12 $\pm$ 3.39 & 66.22 $\pm$ 2.97 \\
 & CLIP-FT & 17.26 $\pm$ 2.07 & 18.25 $\pm$ 4.69 & 66.96 $\pm$ 2.17 & 64.46 $\pm$ 3.20 & 68.70 $\pm$ 2.16 & 67.41 $\pm$ 0.55 & 66.06 $\pm$ 2.54 \\
& FairCLIP\(_{race}\)* & 17.79 $\pm$ 4.86 & 18.30 $\pm$ 2.07 & 69.88 $\pm$ 2.00 & 66.54 $\pm$ 1.73 & 71.78 $\pm$ 2.18 & 71.79 $\pm$ 2.13 & 68.67 $\pm$ 1.99  \\
 & FairCLIP\(_{race}\) & 21.88 $\pm$ 6.11 & 25.07 $\pm$ 8.90 & 64.71 $\pm$ 3.41 & 61.05 $\pm$ 2.16 & 67.01 $\pm$ 4.03 & 66.72 $\pm$ 4.59 & 63.06 $\pm$ 3.26 \\

                       \midrule
&&&&&&\textbf{Female} & \textbf{Male} \\
\multirow{4}{*}{\textbf{Gender}} 
 & CLIP-FT* & 2.93 $\pm$ 3.17 & 4.29 $\pm$ 4.05 & 67.83 $\pm$ 2.92 & 63.86 $\pm$ 2.36 & 65.13 $\pm$ 2.60 & 71.31 $\pm$ 3.24 \\
 & CLIP-FT & 4.17 $\pm$ 1.99 & 8.93 $\pm$ 2.33 & 66.96 $\pm$ 2.17 & 63.08 $\pm$ 1.69 & 64.25 $\pm$ 1.80 & 70.40 $\pm$ 2.74 &   \\
& FairCLIP\(_{gender}\)* & 5.82 $\pm$ 1.22 & 8.14 $\pm$ 2.62 & 69.74 $\pm$ 0.95 & 66.00 $\pm$ 1.55 & 67.29 $\pm$ 1.38 & 72.99 $\pm$ 0.83  \\
 & FairCLIP\(_{gender}\) & 6.14 $\pm$ 1.61 & 8.66 $\pm$ 1.61 & 64.88 $\pm$ 1.05 & 61.87 $\pm$ 0.95 & 62.78 $\pm$ 1.10 & 67.64 $\pm$ 1.24 &   \\

                       \midrule
&&&&&&\textbf{Non-Hispanic} & \textbf{Hispanic} \\
\multirow{4}{*}{\textbf{Ethnicity}} 
 & CLIP-FT* & 7.78 $\pm$ 5.50 & 11.56 $\pm$ 9.03 & 67.83 $\pm$ 2.92 & 62.20 $\pm$ 0.82 & 68.14 $\pm$ 3.07 & 59.13 $\pm$ 0.92
 \\
  & CLIP-FT & 10.56 $\pm$ 3.23 & 16.64 $\pm$ 2.36 & 66.96 $\pm$ 2.17 & 62.16 $\pm$ 1.14 & 67.22 $\pm$ 2.22 & 59.50 $\pm$ 0.71 &   \\
& FairCLIP\(_{ethnicity}\)* & 9.44 $\pm$ 3.28 & 13.03 $\pm$ 5.56 & 69.87 $\pm$ 1.05 & 64.90 $\pm$ 2.05 & 70.16 $\pm$ 1.05 & 62.43 $\pm$ 2.99  \\ 
 & FairCLIP\(_{ethnicity}\)  & 8.96 $\pm$ 5.01 & 12.59 $\pm$ 10.94 & 67.13 $\pm$ 4.15 & {62.25} $\pm$ 3.09 & 67.41 $\pm$ 4.25 & {59.57} $\pm$ 3.60 &   \\

\midrule                        
&&&&&&\textbf{English} & \textbf{Spanish} & \textbf{Others} \\
\multirow{4}{*}{\textbf{Language}} 
 & CLIP-FT* & 7.62 $\pm$ 5.39 & 10.84 $\pm$ 9.84 & 67.83 $\pm$ 2.92 & 61.29 $\pm$ 1.28 & 67.77 $\pm$ 3.18 & 63.83 $\pm$ 3.20 & 62.24 $\pm$ 2.25 \\
 & CLIP-FT & 14.21 $\pm$ 1.45 & 18.76 $\pm$ 3.73 & 66.96 $\pm$ 2.17 & 60.37 $\pm$ 4.01 & 66.92 $\pm$ 2.11 & 63.92 $\pm$ 7.00 & 60.51 $\pm$ 3.67 \\
& FairCLIP\(_{language}\)* & 11.63 $\pm$ 3.53 & 11.05 $\pm$ 5.66 & 68.68 $\pm$ 2.26 & 61.50 $\pm$ 2.29 & 68.71 $\pm$ 2.31 & 65.06 $\pm$ 3.96 & 61.18 $\pm$ 1.20 \\
 & FairCLIP\(_{language}\) & 8.32 $\pm$ 5.71 & 16.53 $\pm$ 10.97 & 68.53 $\pm$ 3.24 & 60.02 $\pm$ 4.14 & 68.69 $\pm$ 3.29 & 63.54 $\pm$ 6.59 & 59.37 $\pm$ 2.49 \\
\bottomrule
\end{tabular}
}
\end{center}
\end{table}
\newpage
\section{Distance analysis for A-FairCLIP}\label{appendix:distance_analysis}
\subsection{Distance analysis for the Sinkhorn distance}
The distances of A-FairCLIP is shown in table~\ref{tab:distances_b16_refactored}. We conclude that when the fairness regularizer rate is sufficiently small, the performance is comparable to fine-tuned CLIP, for both fairness and performance metrics. However, the distances for the model with the small regularizer rate are in most cases comparable with the fine-tuned CLIP distance, and in some cases even larger. Thus, we do not see a fairness rate where the fairness is improved and the performance does not drop significantly.

\begin{table}[h]
\caption{Comparison of Sinkhorn distances per group for different CLIP versions using the ViT-B/16 architecture, based on the diagonal of the similarity matrix (aligned implementation). Note that CLIP has only been run once, since it has pre-defined weights.}
\label{tab:distances_b16_refactored}
\begin{center}
\adjustbox{width=\textwidth}{
\begin{tabular}{llrrrr}
\toprule
\textbf{Attribute} & \textbf{Group} & \textbf{CLIP} & \textbf{CLIP-FT} & \textbf{A-FairCLIP\(_{race}\)}\\ 
\midrule
\multirow{3}{*}{\textbf{Race}} 
    & Asian        & $5.04\cdot 10^{-1}$ &$ 2.16\cdot 10^{-1} $$\pm$ $1.37\cdot 10^{-1}$ & $ 6.44 \cdot 10^{-2}$ $\pm $$ 6.34 \cdot 10^{-2}$\\
    & Black        & $3.62\cdot 10^{-1}$ &$ 8.63\cdot 10^{-2} $$\pm$ $5.60\cdot 10^{-2}$ &  $ 1.37 \cdot 10^{-1}$ $\pm $$ 1.18 \cdot 10^{-1}$\\
    & White        & $4.66\cdot 10^{-2}$   &$ 5.82\cdot 10^{-3} $$\pm$ $3.20\cdot 10^{-3} $ & $ 6.95 \cdot 10^{-3}$ $\pm $$ 5.67 \cdot 10^{-3}$\\ 
\midrule
\multirow{2}{*}{\textbf{Gender}} 
    & Female       & $2.84\cdot 10^{-3}$  &$ 2.55\cdot 10^{-2} $$\pm$ $1.76\cdot 10^{-2}$ & $ 4.90 \cdot 10^{-3}$ $\pm $$ 4.42 \cdot 10^{-3}$\\
    & Male         & $3.95\cdot 10^{-3}$  &$ 6.27\cdot 10^{-2} $$\pm$ $5.53\cdot 10^{-2}$ & $ 7.91 \cdot 10^{-3}$ $\pm $$ 7.50 \cdot 10^{-3}$\\ 
\midrule
\multirow{2}{*}{\textbf{Ethnicity}} 
    & Non-Hispanic & $3.58\cdot 10^{-4}$    &$ 7.53\cdot 10^{-4} $$\pm$ $5.16\cdot 10^{-4}$ & $ 1.22 \cdot 10^{-4}$ $\pm $$ 5.86 \cdot 10^{-5}$\\
    & Hispanic     & $1.65\cdot 10^{-1}$ &$ 3.64\cdot 10^{-1} $$\pm$ $2.17\cdot 10^{-1}$ & $ 3.69 \cdot 10^{-2}$ $\pm $$ 1.38 \cdot 10^{-2}$\\ 
\midrule
\multirow{3}{*}{\textbf{Language}} 
    & English      & $9.30\cdot 10^{-4}$    & $ 1.55\cdot 10^{-3} $ $\pm$ $1.07\cdot 10^{-3}$ & $ 2.01 \cdot 10^{-4}$ $\pm $$ 1.72 \cdot 10^{-4}$\\
    & Spanish      & $2.46\cdot 10^{-1}$ & $1.31 \cdot 10^{0}$~~ $\pm$ $1.04 \cdot 10^{0}$ ~ & $ 1.31 \cdot 10^{-1}$ $\pm $$ 7.99 \cdot 10^{-2}$\\
    & Other        & $2.59\cdot 10^{-1}$ &$ 7.97\cdot 10^{-1} $ $\pm$ $7.62\cdot 10^{-1}$ & $ 3.41 \cdot 10^{-2}$$\pm $$ 3.01 \cdot 10^{-2}$\\ 
\bottomrule
\end{tabular}
}
\end{center}
\end{table}

\subsection{Distance analysis for the MMD distances}
Table~\ref{tab:mmd_gaussian_distances_b16} and Table~\ref{tab:tab:mmd_laplacian_distances_b16} show the distances when using a Gaussian and Laplacian kernel with the MMD.
The models were trained using the aligned code (see Section~\ref{section:code}). The lambda's were chosen based on the hyperparameter tuning process, where the parameters were chosen from the large fairness loss rate interval. This resulted in a learning rate of $1.97 \cdot 10^{6}$ combined with a fairness rate of $200$ for the MMD with a Gaussian kernel, and a learning rate of $3.95 \cdot 10^{7}$ with a fairness rate of $500$ for the MMD with a Laplacian kernel.
Overall, the distances are relatively small for all models (with and without fine-tuning). Since there is no significant difference in distances between the models with and without the FairCLIP objective, we conclude that even with large fairness rates, the FairCLIP objective has little to no effect.

\begin{table}[h]
\caption{Comparison of the MMD distances with a Gaussian kernel per group for different CLIP versions using the ViT-B/16 architecture, based on the diagonal of the similarity matrix. Note that CLIP has only been run once, since it has pre-defined weights.}
\label{tab:mmd_gaussian_distances_b16}
\begin{center}
\adjustbox{width=\textwidth}{
\begin{tabular}{llrrrrrr}
\toprule
\textbf{Attribute} & \textbf{Group} & \textbf{CLIP} & \textbf{CLIP-FT} & \textbf{A-MMD\_{Gaussian}-\text{FairCLIP}\(\_{race}\)} & \textbf{A-MMD\_{Gaussian}-\text{FairCLIP}\(\_{language}\)} \\ 
\midrule
\multirow{3}{*}{\textbf{Race}} 
    & Asian        & $ 3.83 \cdot 10^{-3}$ & $ 2.80 \cdot 10^{-3}$ $\pm$ $ 2.50 \cdot 10^{-4}$ & $ 3.05 \cdot 10^{-3}$ $\pm$ $ 9.24 \cdot 10^{-4}$ & $ 2.42 \cdot 10^{-3}$ $\pm$ $ 5.97 \cdot 10^{-4}$ \\
    & Black        & $ 1.70 \cdot 10^{-3}$ & $ 1.50 \cdot 10^{-3}$ $\pm$ $ 2.14 \cdot 10^{-4}$ & $ 1.40 \cdot 10^{-3}$ $\pm$ $ 2.24 \cdot 10^{-4}$ & $ 1.39 \cdot 10^{-3}$ $\pm$ $ 1.54 \cdot 10^{-4}$ \\
    & White        & $ 9.72 \cdot 10^{-5}$   & $ 8.38 \cdot 10^{-5}$ $\pm$ $ 1.02 \cdot 10^{-5}$ & $ 6.76 \cdot 10^{-5}$ $\pm$ $ 8.93 \cdot 10^{-6}$ & $ 6.52 \cdot 10^{-5}$ $\pm$ $ 8.49 \cdot 10^{-6}$ \\ 
\midrule
\multirow{2}{*}{\textbf{Gender}} 
    & Female       & $ 1.78 \cdot 10^{-4}$  & $ 2.34 \cdot 10^{-4}$ $\pm$ $ 1.02 \cdot 10^{-5}$ & $ 2.18 \cdot 10^{-4}$ $\pm$ $ 2.27 \cdot 10^{-5}$ & $ 2.51 \cdot 10^{-4}$ $\pm$ $ 5.32 \cdot 10^{-5}$ \\
    & Male         & $ 2.64 \cdot 10^{-4}$  & $3.55\cdot 10^4$ $\pm$ $ 1.51 \cdot 10^{-5}$ & $ 3.23 \cdot 10^{-4}$ $\pm$ $ 3.36 \cdot 10^{-5}$ & $ 3.72 \cdot 10^{-4}$ $\pm$ $ 7.90 \cdot 10^{-5}$ \\ 
\midrule
\multirow{2}{*}{\textbf{Ethnicity}} 
    & Non-Hispanic & $ 9.88 \cdot 10^{-6}$    & $ 1.11 \cdot 10^{-5}$ $\pm$ $ 1.11 \cdot 10^{-6}$ & $ 1.16 \cdot 10^{-5}$ $\pm$ $ 1.69 \cdot 10^{-6}$ & $ 1.12 \cdot 10^{-5}$ $\pm$ $ 7.07 \cdot 10^{-7}$ \\
    & Hispanic     & $ 5.59 \cdot 10^{-3}$ & $ 6.30 \cdot 10^{-3}$ $\pm$ $ 6.27 \cdot 10^{-4}$ & $ 6.56 \cdot 10^{-3}$ $\pm$ $ 9.54 \cdot 10^{-4}$ & $ 6.31 \cdot 10^{-3}$ $\pm$ $ 4.00 \cdot 10^{-4}$ \\ 
\midrule
\multirow{3}{*}{\textbf{Language}} 
    & English      & $ 1.95 \cdot 10^{-5}$    & $ 1.68 \cdot 10^{-5}$ $\pm$ $ 1.18 \cdot 10^{-6}$ & $ 1.78 \cdot 10^{-5}$ $\pm$ $ 3.44 \cdot 10^{-6}$ & $ 1.51 \cdot 10^{-5}$ $\pm$ $ 2.34 \cdot 10^{-6}$ \\
    & Spanish      & $ 1.54 \cdot 10^{-2}$ & $ 1.70 \cdot 10^{-2}$ $\pm$ $ 1.64 \cdot 10^{-3}$ & $ 1.34 \cdot 10^{-2}$ $\pm$ $ 1.95 \cdot 10^{-3}$ & $ 1.58 \cdot 10^{-2}$ $\pm$ $ 1.99 \cdot 10^{-3}$ \\
    & Other        & $6.07\cdot 10^{-1}$ & $ 5.15 \cdot 10^{-3}$ $\pm$ $ 7.86 \cdot 10^{-4}$ & $ 5.36 \cdot 10^{-3}$ $\pm$ $ 1.67 \cdot 10^{-3}$ & $ 5.40 \cdot 10^{-3}$ $\pm$ $ 2.28 \cdot 10^{-4}$ \\ 
\bottomrule
\end{tabular}
}
\end{center}
\end{table}

\begin{table}[h]
\caption{Comparison of the MMD distances with a Laplacian kernel per group for different CLIP versions using the ViT-B/16 architecture, based on the diagonal of the similarity matrix. Note that CLIP has only been run once, since it has pre-defined weights.}
\label{tab:tab:mmd_laplacian_distances_b16}
\begin{center}
\adjustbox{width=\textwidth}{
\begin{tabular}{llrrrrrr}
\toprule
\textbf{Attribute} & \textbf{Group} & \textbf{CLIP} & \textbf{CLIP-FT} & \textbf{A-MMD\(_{Laplacian}\)-\text{FairCLIP}\(\_{race}\)} & \textbf{A-MMD\_{Laplacian}-\text{FairCLIP}\(\_{language}\)} \\ 
\midrule
\multirow{3}{*}{\textbf{Race}} 
    & Asian        & $ 3.62 \cdot 10^{-3}$ & $ 2.84 \cdot 10^{-3}$ $\pm$ $ 1.92 \cdot 10^{-4}$ & $ 3.15 \cdot 10^{-3}$ $\pm$ $ 6.30 \cdot 10^{-4}$ & $ 2.94 \cdot 10^{-3}$ $\pm$ $ 4.56 \cdot 10^{-4}$ \\
    & Black        & $ 1.67 \cdot 10^{-3}$ & $ 1.49 \cdot 10^{-3}$ $\pm$ $ 1.59 \cdot 10^{-4}$ & $ 1.54 \cdot 10^{-3}$ $\pm$ $ 1.61 \cdot 10^{-4}$ & $ 1.42 \cdot 10^{-3}$ $\pm$ $ 3.31 \cdot 10^{-4}$ \\
    & White        & $ 9.22 \cdot 10^{-5}$   & $ 1.02 \cdot 10^{-4}$ $\pm$ $ 1.59 \cdot 10^{-5}$ & $ 6.76 \cdot 10^{-5}$ $\pm$ $ 8.93 \cdot 10^{-6}$ & $ 7.72 \cdot 10^{-5}$ $\pm$ $ 1.06 \cdot 10^{-5}$ \\ 
\midrule
\multirow{2}{*}{\textbf{Gender}} 
    & Female       & $ 1.90 \cdot 10^{-4}$  & $ 2.28 \cdot 10^{-4}$ $\pm$ $ 6.20 \cdot 10^{-6}$ & $ 2.38 \cdot 10^{-4}$ $\pm$ $ 5.76 \cdot 10^{-5}$ & $ 2.42 \cdot 10^{-4}$ $\pm$ $ 6.81 \cdot 10^{-5}$ \\
    & Male         & $ 2.82 \cdot 10^{-4}$  & $ 3.38 \cdot 10^{-4}$ $\pm$ $ 9.14 \cdot 10^{-6}$ & $ 3.53 \cdot 10^{-4}$ $\pm$ $ 8.55 \cdot 10^{-5}$ & $ 3.59 \cdot 10^{-4}$ $\pm$ $ 1.01 \cdot 10^{-4}$ \\ 
\midrule
\multirow{2}{*}{\textbf{Ethnicity}} 
    & Non-Hispanic & $ 1.07 \cdot 10^{-5}$    & $ 1.11 \cdot 10^{-5}$ $\pm$ $ 7.62 \cdot 10^{-7}$ & $ 1.06 \cdot 10^{-5}$ $\pm$ $ 1.57 \cdot 10^{-6}$ & $ 7.95 \cdot 10^{-6}$ $\pm$ $ 2.34 \cdot 10^{-6}$ \\
    & Hispanic     & $ 6.07 \cdot 10^{-3}$ & $ 6.27 \cdot 10^{-3}$ $\pm$ $ 4.30 \cdot 10^{-4}$ & $ 5.98 \cdot 10^{-3}$ $\pm$ $ 8.87 \cdot 10^{-4}$ & $ 4.49 \cdot 10^{-3}$ $\pm$ $ 1.32 \cdot 10^{-3}$ \\ 
\midrule
\multirow{3}{*}{\textbf{Language}} 
    & English      & $ 1.88 \cdot 10^{-5}$    & $ 1.67 \cdot 10^{-5}$ $\pm$ $ 1.06 \cdot 10^{-6}$ & $ 1.89 \cdot 10^{-5}$ $\pm$ $ 4.56 \cdot 10^{-6}$ & $ 1.29 \cdot 10^{-5}$ $\pm$ $ 4.41 \cdot 10^{-6}$ \\
    & Spanish      & $ 1.61 \cdot 10^{-2}$ & $ 1.68 \cdot 10^{-2}$ $\pm$ $ 1.43 \cdot 10^{-3}$ & $ 2.10 \cdot 10^{-2}$ $\pm$ $ 5.57 \cdot 10^{-3}$ & $ 1.37 \cdot 10^{-2}$ $\pm$ $ 1.98 \cdot 10^{-3}$ \\
    & Other        & $5.78\cdot 10^{-3}$ & $ 5.18 \cdot 10^{-3}$ $\pm$ $ 5.55 \cdot 10^{-4}$ & $ 4.71 \cdot 10^{-3}$ $\pm$ $ 1.18 \cdot 10^{-3}$ & $ 4.38 \cdot 10^{-3}$ $\pm$ $ 1.38 \cdot 10^{-3}$ \\ 
\bottomrule
\end{tabular}
}
\end{center}
\end{table}

\clearpage
\section{Detailed Computational Resources}\label{appendix:resources}
The experiments were carried out on a private infrastructure with an estimated carbon efficiency of 0.270 kgCO\(_2\)eq/kWh~\citep{co2eqkwh_NL} and an estimated power consumption efficiency (PUE) of 1.2~\citep{surfPUE}. The hardware used included a single NVIDIA A100 40GB GPU, with a TDP of 250W, along with 9 Intel Xeon Platinum 8360Y CPUs.

A total of approximately 990 hours of computation were performed on the A100 GPU, resulting in an estimated 80.3 kgCO\(_2\)eq emissions. These emissions were calculated using the  Machine Learning Impact calculator\footnote{\url{https://mlco2.github.io/impact\#compute}} as presented in \citet{lacoste2019quantifying}. 100 kgCO\(_2\)eq were manually offset through Trees For All\footnote{\url{https://treesforall.nl/en/offset-carbon-emissions/}}.

\begin{table}[h]
\caption{Detailed summary of the computational requirements for each experiment including averaged runtime in hours and estimated \(\mathrm{CO}_2\)-emission (kgCO\(_2\)eq).
``Other'' experiments include code testing and experiments not directly related to the main study.}
\label{tab:experiment-summary}
\begin{center}
\adjustbox{width=\textwidth}{
\begin{tabular}{llrrrrrr}
\toprule
\textbf{Experiment Name} & \textbf{Data set} & \textbf{Runs}  & \textbf{Runtime} & \textbf{Total Runtime} &  \textbf{kgCO\(_2\)eq} \\ 
\midrule
Fine-tuning CLIP\(_{ViT-B/16}\)                 & Harvard-FairVLMed & 3  & 0.08  & 0.24  & 0.02 \\
Fine-tuning CLIP\(_{ViT-L/14}\)                 & Harvard-FairVLMed & 3  & 0.23  & 0.69  & 0.06 \\ 
Fine-tuning FairCLIP\(_{ViT-B/16}\)             & Harvard-FairVLMed & 12 & 0.22  & 2.64  & 0.22 \\ 
Fine-tuning FairCLIP\(_{ViT-L/14}\)             & Harvard-FairVLMed & 12 & 0.45  & 5.40  & 0.44 \\ 
Fine-tuning FairCLIP+\(_{ViT-B/16}\)            & Harvard-FairVLMed & 3  & 0.52  & 1.56  & 0.13 \\ 
Fine-tuning FairCLIP+\(_{ViT-L/14}\)            & Harvard-FairVLMed & 3  & 0.75  & 2.25  & 0.19 \\ 
Linear Probing CLIP\(_{ViT-L/14}\)              & Harvard-FairVLMed & 3  & 4.00  & 12.00 & 0.98 \\ 
Linear Probing CLIP-FT\(_{ViT-L/14}\)           & Harvard-FairVLMed & 3  & 4.00  & 12.00 & 0.98 \\ 
Evaluation of all models                        & Harvard-FairVLMed & 16 & 0.03  & 0.48  & 0.04 \\ 
Lambda Fair Loss Search                         & Harvard-FairVLMed & 1  & 23.00 & 23.00 & 1.87 \\ 
Hyperparameter Tuning FairCLIP\(_{ViT-B/16}\)   & Harvard-FairVLMed & 36 & 24.00 & 864.00 & 69.99 \\
Fine-tuning CLIP\(_{RS50}\)                     & FairFace          & 3  & 0.17  & 0.51  & 0.05 \\
Fine-tuning FairCLIP+\(_{RS50}\)                & FairFace          & 3  & 2.00  & 6.00  & 0.49 \\ 
Fine-tuning FairCLIP+\(_{race,RS50}\)           & FairFace          & 3  & 2.00  & 6.00  & 0.49 \\
Other                                           & Both              & -  & -     & 53.23 & 4.32 \\
\midrule
\textbf{Total}                                  & \textbf{-}        & \textbf{-}   & \textbf{-}    & \textbf{990.00}   & \textbf{80.30}   \\
\bottomrule
\end{tabular}
}
\end{center}
\end{table}

Table \ref{tab:experiment-summary} summarizes the computational requirements for each experiment to ensure transparency by detailing the number of runs, average runtime, total runtime, and estimated CO\(_2\) emissions for each experiment.

\end{document}